# Spatial and Temporal Characteristics of Freight Tours: A Data-Driven Exploratory Analysis

Ali Nadi[a*]

Lóránt Tavasszy[a]

J.W.C. van Lint[a]

Maaike Snelder[a]

[a] Delft University of Technology, Delft, The Netherlands

* Corresponding author: a.nadinajafabadi@tudelft.nl
2023

## Abstract

This paper presents a modeling approach to infer scheduling and routing patterns from digital freight transport activity data for different freight markets. We provide a complete modeling framework including a new discrete-continuous decision tree approach for extracting rules from the freight transport data. We apply these models to collected tour data for the Netherlands to understand departure time patterns and tour strategies, also allowing us to evaluate the effectiveness of the proposed algorithm. We find that spatial and temporal characteristics are important to capture types of tours and time-of-day patterns of freight activities. Also, the empirical evidence indicates that carriers in most of the transport markets are sensitive to the level of congestion. Many of them adjust the type of tour, departure time, and the number of stops per tour when facing a congested zone. The results can be used by practitioners to get more grip on transport markets and develop freight and traffic management measures.

## 1 Introduction

Two key structural characteristics of tours in freight transportation are schedules for pick-up/delivery and the routing patterns connecting the stops. Identifying the structure of tours is useful for freight transportation decision-makers and traffic managers, as they support the descriptive and predictive analysis of the freight transportation system. It is important for freight and traffic management, as these structures are conditioned upon and also determine the state of the traffic on motorways. From the demand management perspective, modeling time-of-day choices casts light on the scheduling of departures and modeling tours elucidates routing patterns of commercial vehicles. As opposed to passenger travel, researchers have devoted relatively little attention to modeling freight transport trips. The reason for this is the complexity of this part of the traffic system, particularly as freight transportation includes many submarkets (Khan and Machemehl, 2017a, Holguín-Veras and Patil, 2005). Previous research has addressed multiple dimensions of freight trips (Garrido and Mahmassani (2000), Nuzzolo and Comi (2014), Wisetjindawat et al. (2012), and Figliozzi (2010)). To the best of our knowledge, however, the structure of tours for different submarkets, the interactions between industries, and the sensitivity to traffic conditions have not yet been analyzed in detail. This research aims to help fill this gap by developing a series of models that help explain the mechanisms, based on disaggregate shipment and trip observations.

The opportunity for this work arose from access to a large and rich database of micro-level observations of freight transport activities in the Netherlands. As the main direction of the work, we consider three dimensions of tour characteristics i.e. time, type, and size. These entail departure times of tours, the sequence of loading and unloading activities, and tour length. These decisions are often the outcome of complex and interdependent routing and scheduling optimization processes (OR tools). In other words, route planners mostly follow tours recommended by tools like Google OR or optimization software like Gurobi. Although unique tours will typically be suggested by optimization packages on an individual case basis, we believe that viewed across a larger population of firms, the tours resulting from these tools will display similarities under appropriate circumstances. Knowledge of these similarity patterns is valuable for practitioners to know and exploit. Here, we aim to distill these patterns using a machine learning approach. In our study, we are interested to identify patterns and the circumstances in which they occur from a given dataset, without enforcing any theory or behavioural assumptions on this process ex-ante.

In this paper, we propose a machine learning approach to describe the structure and timing of freight tours, based on association rule mining and decision trees. We use rule mining to investigate relationships between different clusters of freight transport markets, and decision trees to discover tour and timing patterns from a large database of freight transport activities. We use the information extracted from 9 identified transport markets. Specifically, we ask the following questions:

1.  How do tour strategies and external circumstances affect the peak-hour occurrence of tours?

2.  How do the types of logistic nodes (e.g. distribution centers, transshipment terminals) influence departure times and tour structures?

3.  How do transport costs influence the structure of tours?

4.  How does the number of shipments (i.e. from empty containers to containers with many shipments) affect the scheduling of tours?

5.  How does the type of vehicle affect the routing and scheduling of shipments?

These research questions are answered using a two-step approach. In the first step, we use decision trees to explore patterns in departure times of tours. Earlier studies have identified that freight tours encompass multiple inter-related decision variables, such as the type of tour pattern and the number of stops in each selected tour type (Khan and Machemehl, 2017a). In the second step, we mine probabilistic rules that (a) describe the various pick-up and delivery strategies and (b) predict the average number of stops per tour for each strategy simultaneously. Prediction of a mixture of target variables is needed in the second modeling task. However, most machine learning techniques can only predict one type of target variable at a time. To deal with this problem, we introduce an enhanced decision tree algorithm that can predict joint mixed target variables.

These two models together explain the anatomy of tours taking both scheduling and routing activity of various transport markets into account. The main contributions of this paper are the following:

1.  We introduce a new data processing framework to support the process of inferring patterns in big, disaggregate freight tour databases.

2.  We extract rules from observations to cluster the freight transport market and investigate the interaction of multiple industries, through an analysis of their tour activities.

3.  We propose a new decision tree algorithm that predicts mixed discrete and continuous target variables to classify tour structures and predict the number of stops per tour strategy. This innovative algorithm can be used for modeling similar cases with other mixed-type target variables.

4.  We provide new insights into how transport companies plan their tours in the presence of congestion.

The paper is organized as follows: Section 2 provides a brief overview of the existing studies on the topic of this paper. Section 3 presents the methodology with which we characterize freight activities, including the description of the data structure and the fusion of multiple data sources. Section 4 presents the findings from the descriptive analysis of the freight routing and scheduling patterns. Section 5 sets up a discussion about findings, provides validation against literature, and suggests directions for further research. Section 6 offers concluding insights.

## 2    Literature review

This section presents a brief review of the literature that concerns freight tour activity modeling. The literature reports several important factors to take into account. Primarily, it is the operational decisions of shippers, senders, receivers, and carriers that determine truck flows (Khan and Machemehl, 2017a). Logistics hubs (e.g. distribution centers and transshipment terminals) may add complexity to the tours due to their different functionalities for storage and cross-docking of freight. Tour generation is dependent on important dimensions that are often difficult to measure due to privacy issues. Examples are vehicle type, weight, capacity, container type and dimension, departure and arrival times, number of stops, and tour duration (Khan and Machemehl, 2017a). Besides, freight transportation demand is highly variable over time and space (Garrido and Mahmassani (2000)). The heterogeneity in transport markets and industrial sectors creates additional diversity in tour characteristics.

To cover the above issues, disaggregate activity-based modeling has been developed in the literature capturing daily activity patterns of goods movements. Recently, more research is focusing on agent-based micro-simulation of multiple actors' decisions in freight transport systems. Examples are MASS-GT which is an agent-based simulation system for urban goods transport. MASS-GT simulates the choices of suppliers and receivers, transport service providers, and carriers (de Bok and Tavasszy, 2018). These choices include distribution channel selection, carrier selection, vehicle type, shipment size, and routing and scheduling decisions. Another example is SimMobility developed by Sakai (Sakai et al., 2020) which can simulate disaggregate interactions of multiple agents that are engaged in freight-related activities. These decisions include commodity contracts, overnight parking place choice, carrier selection, and vehicle operation choices. Lately, agent-based freight simulators are also integrated with agent-based passenger simulators like MATSIM to model the impact of freight transport decisions on traffic system (Schröder and Liedtke, 2017). For example, Mommens et al. (2018) used TRABAM, which is a freight simulator linked to MATSIM, for the impact assessment of traffic management scenarios like night distribution.

All in all, the common choice models that exist in most agent-based freight simulators include (a) tour purpose and vehicle type choice; (b) next stop destination choice; (c) next stop purpose choice; (d) joint tour-type and number-of-trip choice and (e) departure time choice. Among all, time-of-day (i.e. scheduling) and type-of-tour (i.e. routing) are the two main characteristics of freight transport activities that are believed to be conditioned upon the level of congestion on motorways (Khan and Machemehl, 2017b) and thus are of interest for traffic and transport managers.

Time-of-day modeling is relevant as variations in departure time of freight flows may have large impacts on motorway congestion; whereas in turn, those congestion problems may also heavily affect logistics operations (Figliozzi, 2010). The application of a time-of-day model helps researchers to evaluate shifts between peak and off-peak periods for freight transport as a means to avoid congestion on roads. In spite of its significance, there exists little research on freight transport departure time policies. The earliest time-of-day modeling frameworks utilized Monte Carlo simulation in which departure times were averaged from a sample of (limited) observations (Hunt and Stefan, 2007). Probably, the study of Nuzzolo and Comi (2014) is one of the first studies that modeled the departure time as a part of the freight demand modeling framework. In this study, like most similar studies, a discrete choice model was estimated on urban freight trip survey data to calculate the probability of a delivery tour departure time from an origin. The results indicate that the departure time of a tour is strictly correlated with the number of stops per tour. A recent time-period choice model based on stated preferences for road freight transport can be found in (De Jong et al., 2016). They estimated a multinomial logit model on SP data of 158 receivers of the goods. This model captures the sensitivity of peak and off-peak delivery choices to changes in travel time and cost. It shows a stronger sensitivity to travel costs than to travel time. This study was followed up by Vegelien and Dugundji (2018) who used revealed preference data (GPS tracking data of trailers) to model time-of-day choice, with a nested logit model. This model includes trip duration and product type as explanatory variables. Khan and Machemehl (2017b) proposed a discrete-continuous probit model to recognize trucks' time-of-day preferences and predict the vehicle-mile traveled at a specific time of the day. This model was estimated on commercial vehicle survey data. It shows the contribution of spatial and temporal factors including vehicle type, commodity type, unloading weight, characteristics of intermediate stops speed, and service time. However, this study does not include the impact of traffic on time-of-day preferences which has our interest.

Similar to the time-of-day model, a small number of studies specifically investigate tours. An early study of commercial vehicle delivery strategy can be found in (Burns et al., 1985). They formulate transportation by truck and inventory costs of freight to evaluate the cost trade-off between the direct shipping and collection or distribution type of tour strategies, based on shipment size and the number of customers. They show that in the direct shipping strategy the optimal shipment size follows the economic order quantity (EOQ) rule while in collection or distribution, the shipment size is close to a full truckload. Ruan et al. (2012) followed this study further using mixed and multinomial logit models to model multiple types of tours. This model helps to understand decisions regarding distribution strategies and tour chaining. Zhou et al. (2014) studied tour patterns of urban commercial vehicles based on the number of trips made for delivery and pickup activities. They defined four tour-type alternatives bundling different ranges of the number of trips with different tour types. Then, they used multinomial and nested logit models to identify daily trip chaining behavior. Khan and Machemehl (2017a) considered the number of trips as a continuous dimension of tour activities. This is more appropriate since it refers to the size of tours and there is ordinal relation between tours of different sizes. They develop a multiple discrete-continuous choice method to model a joint distribution of the type of tour strategy and the number of trips for commercial vehicles. Based on their findings, the number of trips and type of tour are two inter-related joint decisions and unified discrete-continuous modeling is a more appropriate modeling framework to capture commercial vehicle movement patterns. The results of this study also indicate the impact of trip features, commodity characteristics, and attributes of base and intermediate stop locations on choice of the type of tour and the number of stops per each tour chain strategy. Most recently Siripirote et al. (2020) proposed a statistical approach to estimate truck activities, commodity-related trip chains, and the status of legs (e.g. empty, full, or partially loaded trips) from collected GPS data. This class of demand modeling uses different units of analysis, either trips, as in Hunt and Stefan (2007) or shipments (as in Thoen et al. (2020) and Nuzzolo and Comi (2014)) and utilizes choice models to explain daily tour patterns.

In summary, the mentioned studies underpin the importance of modeling the time-of-day and type of tour strategies of commercial vehicles in urban areas. We note several gaps, however. Firstly, to the best of our knowledge, existing studies on time-of-day and type of tour analysis relate to urban logistics and do not consider freight transport activity between large suppliers, manufacturers, and logistics hubs. Secondly, Most studies in the field are built upon a limited sample of survey data from the sender or receiver of goods. The literature, therefore, lacks a systematic data-driven modeling pipeline that allows us to estimate models on large-scale real-world tour data derived from executed tours. Finally, segmentation of transport markets based on the interaction between multiple industries and their impact on time and type of tour analysis has not yet been presented in the literature.

To help fill these gaps in research, we develop a new modeling approach utilizing and enhancing rule-based machine learning techniques to explore the departure time, type of tour, and the number of stops per tour pattern, using an extensive freight transport tour database. The next section elaborates on the methodology of the research.

## 3    Modeling framework for knowledge extraction from tour data

In this section, we explain the methods and data used for the analysis of the freight tour databases. Figure 1 shows the overall rule-based modeling framework that allows us to extract appropriate rules about departure time and type of tour strategies in freight transport activities. This framework includes 3 main steps from the preprocessing of data to the full analysis of the structure of tours.

1.  In the first step, data is collected from databases that include the trip survey data and contextual data relating to speeds as well as the locations of important hubs including distribution centers and transshipment terminals.

2.  In the preprocessing step, data is prepared through the fusion of databases and clustering of transport markets. Also, the proximity to congestion zones is modeled to prepare for the tour structure analysis.

3.  In this third step, an enhanced decision tree algorithm is used to model time-of-day patterns, type-of-tour strategies, and the number of stops.

The next subsections 3.1-3.3 describe the details of the approach in this order.

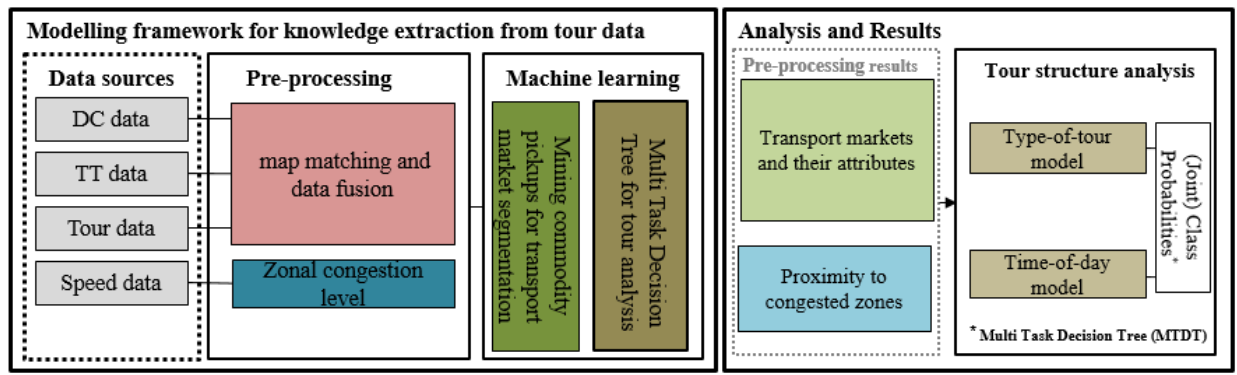

**Figure 1: Procedure for modeling and extracting knowledge from freight tour data**

### 3.1    Data sources

In this study, We make use of a unique and detailed dataset of truck diaries provided by the Centraal Bureau voor de Statistiek (CBS), or Statistics Netherlands. In total CBS provides a set of digitally collected 2.7 million records of data related to freight tour activities from the year 2015. For each commodity in a tour, there is a record in this dataset that contains geographic information regarding (un)loading locations, commodity type, vehicle types used, and other tour characteristics. The data however lacks logistic activity of the visiting locations or the sender or receiver of the shipments. We require this information to understand how different types of logistics activities may change the structure of tours. To enrich the database, we fused it with two external data sources. The first dataset contains over 1600 registered distribution centers along with their 6-digit postcode, size, and sectors. This dataset was sourced from Directorate-General for Public Works and Water Management (Rijkswaterstaat) in the Netherlands. The second dataset contains transshipment terminals (TT) from the IDVV-binnenvaart game (IDVV inland shipping game) containing information regarding 76 terminals and their postcode, size, and annual throughputs. Finally, we used national firm establishment data that includes firms' registration dates, sizes, and business types. More information about these data and their use of it can be found in Zhang and Tavasszy (2012). As described in the next section, we used a matching algorithm to match the pickup and delivery location of vehicles to the right logistic activities using these two data sources.

## 3.2    Pre-processing

### 3.2.1    Map matching and data fusion

To impute the activity type of loading and unloading locations in CBS tour data we propose a hierarchical assignment approach to national geographical divisions based on both 6 (PC6) and 4 (PC4) digits postcodes. Having L ={$l_1,l_2,...,l_n$} as a set of all geographical locations, we can define $L_{6-digits} \subset L_{4-digits} \subset L$ where $L_{6-digit}$ has the highest resolution in national geographical divisions. We give precedence to the PC6 because the divisions are smaller and thus the assignment takes place with higher certainty. We then aggregate the information at this level for use in PC4 where the imputations are more probabilistic.

The algorithm, first, uses firms establishment data and available national statistics to calculate the probability $P_{s,c}^{a,l}$. This shows the probability that a shipment $s$ that picks up or delivers commodity type $c$ at its loading or unloading location $l$ belongs to a firm with activity type $a$.

$$P_{s,c}^{a,l_{6-digit}} = \frac{N_{a,l^{6-digit}} \times P_{a,c}^{make/use}}{\sum_{i \in A} N_{i,l^{6-digit}} \times P_{i,c}^{make/use}} \qquad \forall s \tag{1}$$

Where $N_{a,l}$ is the number of firms with any of the activity types A={ distribution centers, transshipment terminal, producer, consumer } in each location $l$ and $P_{a,c}^{make/use}$ denotes the probability that firms with activity type $a$ are making (if $l$ is loading location) or using (if $l$ is unloading location) of the commodity type $c$. Then the algorithm uses the Monte Carlo simulation to assign each activity type to the loading and unloading locations of a shipment based on $P_{s,c}^{a,l}$.

Please note that in cases that the loading and unloading location of a shipment is reported in a higher resolution i.e. 6-digit, the $P_{s,c}^{a,l}$ is in many cases 1 or close to one as there is a lower chance of having multiple firms with different activity types in a 6-digit zip code. Therefore, the assignment process in these cases is more deterministic than probabilistic. However, this is the opposite when the reported locations are in 4-digits and the assignments would have a high level of uncertainty. To reduce this uncertainty level for 4-digit locations, the algorithm uses the results from assignments in 6-digit resolution. First, the probability that a shipment with commodity type $c$ goes to a zone $l_{6-digit}$ inside the reported larger geographical division (4-digit) is calculated.

$$P_{c,l_{6-digit}} = \frac{M_{c,l_{6-digit}}^{in/out}}{\sum_{l \in L_{6-digit} \subset l_{4-digit}} M_{c,l}^{in/out}} \quad \forall l_{4-digit} \tag{2}$$

Where $M_{c,l}^{in/out}$ is the number of shipments with commodity type c that go inside or come out of the zone $l$. Then a Monte Carlo simulation is used to select and assign each shipment in a tour to an $l_{6-digit}$ location inside the reported $l_{4-digit}$. Afterward, a similar approach as in 6-digit postcode zones (i.e. equation 1) is used to assign an activity type to the visited location.

There are cases where the commodity type is missing or not reported in the dataset. However, our algorithm calculates the assignment probabilities without considering the term $P_{a,c}^{make/use}$ and indices $c$ in equations 1 and 2. In other words, the assignment probabilities only depend on $N_{a,l}$. Please note that these cases are not considered for further analysis in this research.

### 3.2.2 Zonal congestion level

Besides logistics data, we also required traffic data to understand what the traffic conditions were at the shipment time. Below, we explain the methods used to obtain traffic conditions by making use of traffic speed data.

Carriers usually use planning software to schedule the tours. This software uses the average travel time while making routing and scheduling tables. To understand how the congestion can affect tour characteristics, we require a perception of congestion at the time of deliveries. We make use of vehicle speed data derived from loop detectors to obtain congestion information for the pickup and delivery locations. We use the same method introduced by (Christidis and Rivas, 2012) to calculate the aggregate congestion level for each geographical zone. In this approach, a moving average of delays is determined as the congestion indicator. To calculate this delay, we first calculate the moving average speed of each loop detector on a road over a time filter T to smooth very short-term fluctuations. Note that the speed in Equation (3) is instantaneous.

$$\bar{v_i} = \frac{1}{T} \sum_{j=i}^{j+T} v_j \tag{3}$$

Where $i$ is the start of the period, and $v_j$ is the average speed obtained from sensors on the link for the time step $j$. While we define $v_{min}$ as the lowest average speed for the segment, the free flow speed $v_{free}$ is the maximum measured average speed:

$$v_{min} = \min_{j \in [t_0, t]} \bar{v}_j \tag{4}$$

Where $t_0$ and $t$ are the start and end of a period $P$. Then the average delay during period $P$ for each segment is:

$$d_P = \frac{60}{(\frac{1}{v_{min}^p} - \frac{1}{v_{free}^p})} \tag{5}$$

We use $d_P$ as an indicator to define the congestion level for period $P$ measured in minutes per kilometer. One can specify the $P$ parameter to limit the delay estimation for a specific period. We set $P = 1$ hour for each departure time specification i.e. morning peak, midday, evening peak hour, and rest of the day. Based on the average delay indicator, We calculate the 1-hour aggregate of the congestion level (CL) for every zone (zoning system is based on the 4 digits postcode):

$$CL_{z,P} = \frac{\sum_{r \in z} l_r d_{r,p}}{\sum_{r \in z} l_r} \tag{6}$$

Where $l_r$ is the length of segment $r$ in zone $z$. We consider 10 seconds delay for every kilometer in the peak period, as recommended in (Christidis and Rivas, 2012), to identify if a zone is congested.

$$CI_{z,P} = \begin{cases} 1 & CL_{z,P} > 10s \\ 0 & otherwise \end{cases} \tag{7}$$

Where $CI_{z,P}$ is the congestion indicator which is 1 if the zone is congested and 0 otherwise. In section 4.1, we explain how this indicator can come into use for analyzing tour structure.

### 3.3 Machine Learning

#### 3.3.1 Mining Commodity pickups for transport market segmentation

Clustering the activity of carriers is an important preliminary step in understanding disaggregate patterns of tour structures, due to the heterogeneity of decision rules and conditions in different transport markets. In this section, we adopt a machine learning method that is used in the analysis of structured data collected from a sequence of activities of individuals such as in market basket analysis (Raeder and Chawla, 2011). We use this method for clustering transport markets considering their tour activities. To the best of our knowledge, this method, which we adapt in several ways for our purposes, has not been applied to transportation.

The Classification of Products by Activity (i.e. CPA-classification) is an international standard that consists of a complete hierarchy of divisions, groups, and classes. This standard classifies companies based on their economic activity. The CPA-classification enables governments to describe companies according to the type of product that they produce. Since 1967, the standard goods classification for transport statistics (NST-R) has been used and linked to the CPA-classification at the European level (Liedtke and Schepperle, 2004). The NST-2007, for example, consists of 81 goods categories which are aggregated under 20 main commodity type chapters. Although NST-R gives a direct impression to the economic sectors, it is not clear how carriers can work with different sectors. One carrier may just adopt his utilities to carry agricultural products while others may use other facilities to transport mineral products. These carriers may have different decision rules in planning their tours due to the sectors' conditions and protocols. However, some carriers work with multiple sectors. This happens when there is an interaction between two or more sectors in the supply chain. Therefore, CPA-classification alone cannot be used to cluster transport markets accurately. To understand accurately how carriers' tour planning rules differ, we need to cluster them based on the different sectors that they work with. In this section, we use an association rule mining technique to introduce a commodity pickup analysis. This analysis is to extract a set of frequent rules in pickup and delivery data to explain how carriers interact with multiple sectors. In this analysis, we have a set of Tours T={$T_1$,$T_2$,..,$T_n$} in the CBS freight transport diary database. Having a set of commodity items I={$I_1$,$I_2$,…,$I_3$} which are the goods categories in NST-2007, each tour $T_i$ consists of a set of pickups with commodity types $I_i$ (i.e. $T_i \subseteq I$). We aim to extract strong rules like $A \implies B$ which implies that carriers who transport commodity $A \subset I$ are also willing to transport commodity $B \subset I$ in their planned tours ($A$ and $B$ are two disjoint subsets where $A \neq \phi, B \neq \phi$, and $A \cap B = \phi$). By definition, a rule is strong if the usefulness and certainty of that rule are higher than minimum thresholds. We can measure the usefulness and certainty of a rule using *Support* and *Confidence* indicators respectively.

$$Support\left(A \to B\right) = P\left(A \bigcap B\right) \tag{8}$$

$$Confidence(A \to B) = P(B \mid A) = \frac{P(A \bigcap B)}{P(A)} \tag{9}$$

Where *Support* is the probability that a carrier transports both commodity types A and B, and *Confidence* is the probability that a carrier works with sector B given that they also work with sector A. Although *Confidence* and *Support* can identify strong rules in data, some strong rules may happen randomly and thus we should not necessarily consider them important rules. Therefore, another indicator that can capture the correlation between picking up commodity $A$ and picking up $B$ should be considered. This indicator is known as *Lift*.

$$lift(A \rightarrow B) = \frac{P(A \bigcap B)}{P(A) \cdot P(B)} \tag{10}$$

The *Lift* indicator can get any value greater, equal, or less than 1. *Lift* equal to 1 means that the rule happened randomly while *Lift* greater than one implies that the occurrence of picking up commodities A and B has a higher chance than just a random occurrence. In other words, the *Lift* indicator considers the correlation between the occurrence of picking up commodity types A and B. Therefore, the greater the *Lift* is, the higher chance the rule has of occurring.

The complexity of finding all strong and important rules is $2^m - 1$ for m items in the item set. This means that verifying all the possible rules in the database is, in some cases, impossible. However, algorithms like Apriori (Agrawal and Srikant, 1994) use the close-set and maximal-set concepts to limit the search space and solve such complex problems. This algorithm iteratively uses a sequence of pruning and joining processes to capture the most prevalent rules with minimum *Support* and *Confidence*. Based on the most important rules derived from this method, we can understand how carriers interact between different sectors and we can cluster transport markets into a number of submarkets with similar pickup and delivery patterns. We use the 'arule' package in the R programing language (Hornik et al., 2005, Hahsler and Chelluboina, 2011) for implementation and discuss the result of this analysis in section 4.1.2.

### 3.3.2   Multi-task decision tree for tour analysis

Given all previous pre-processing steps, we can now take a step forward towards the development of a method for explaining the anatomy of tours for each of the submarkets. Several machine learning techniques can be used to classify or predict the characteristics of tours. There is a non-linear relation between feature space and dependent variables in this context. Therefore, we need a nonlinear method like kernel-based support vector machines or neural networks that can deal with such non-linearity. However, these methods usually suffer from a lack of interpretability. Among other machine learning techniques, there are two classes of methods that can deal simultaneously with non-linearity and interpretability. The first class is the generalized additive model (GAM) that was introduced by Hastie (1987). GAM is a sum of weighted smooth functions that apply transformations in the linear regression models. The focus of parameters in this approach is on the inference made by these smooth functions on the predictors. The second class is the Decision trees (DT) model which is a supervised learning method with a wide range of applications in both regression and classification problems (Breiman et al., 1984). This method obtains a set of rules by taking the interaction of covariates and can explain the variability of the response variable by recursive partitioning of all the data according to the most significant covariate (Loh, 2014). In other words, a decision tree is a graphical representation of all paths to a decision under certain conditions. These models are not only simple but also powerful to build descriptive models. They usually outperform other classic statistical models like linear regression (in regression problems) and logistic regression (in classification problems) if the relationship between covariates and response variables is highly non-linear. Compared to other black boxes non-linear machine learning predictors with high prediction accuracy, DT is fully explainable and descriptive. The other advantage of the Decision Tree method is that it makes no assumptions regarding the probability distribution of variables (as we do for instance in naive Bayesian classifiers) but is still able to identify explanatory variables and detect interactions among them. All these reasons made us choose to employ and enhance this methodology for our use. In the next section, we explain how and why this method is enhanced to be used for the structure of tour analysis.

Most machine learning methods including decision trees usually support single target variables. However, In some use cases, it is important to model multiple target variables simultaneously. In the modeling terminology, if the target variables are categorical, then it is called multi-label or multi-target classification, and if the target variables are numerical, then we call it multi-target regression (Borchani et al., 2015). Here, however, we want to construct a decision tree that can predict mixed discrete and integer target variables. The application of this problem is to predict the type of tour strategy (a categorical quantity), and its associated number of stops (an integer quantity) that take place within that strategy. There are some techniques to handle this problem:

1) create multiple single models each for one of the target variables. This method however does not consider the interaction and possible correlation between target variables.

2) One can also create multiple single models each for one target variable with other target variables appended to the feature space. Although this method takes the interaction of target variables into account, there are some drawbacks. First, creating multiple models is difficult to interpret. Besides, in most of the statistical and machine learning models, we assume the attributes are independent. Therefore, putting target variables into the feature space may contradict this assumption. Also, this method can be interpreted as one target variable being conditioned on the others. It cannot be interpreted as a joint occurrence of them which is desired in our case.

3) The third approach builds a single classification model that classes are a pairwise combination of the categorical variable with each level of an integer variable. This increase the number of class in the model and the model may not be tractable. A solution for this is to categorize the integer variable into a few classes. However, finding the right number of clusters for integer variables is still challenging.

4) using a multivariate distribution like Copula is another approach that in the case of integer-discrete variables may lead to a large number of classes with an imbalanced distribution which is challenging for most classification algorithms to search in such feature space.

To handle the mentioned problems, we propose a new approach to build one single (unified) DT model with both integer and discrete target variables. In this case, a DT has to learn multiple tasks, i.e. regression and classification, simultaneously. Therefore we called this algorithm a Multi-Task Decision Tree (MTDT). This approach neither requires discretizing target variables nor makes a large number of classes. It also does not make any assumption about the distribution of target variables and yet can take the dependency between them into account. To this end, we adopt the idea from the ID4.5 algorithm proposed by Quinlan (2014) to introduce a new version of this algorithm, that can handle multiple mixed integers and categorical target variables. We applied this algorithm to model the type-of-tour and the number of stops simultaneously. The ID4.5 algorithm starts splitting from the root node and continues on further nodes. The first step in a decision tree is to decide where to split. In other words, which feature should be used to split the data? For classification purposes, information gain (IG) is used to decide on a cut-point. Information gain is the measure of impurity or randomness and is the decrease in Shannon's entropy after the data set is split based on an attribute.

$$H(C) = -\sum_i p(C = c_i) \log(p(C = c_i)) \tag{11}$$

$$H(C \mid X) = -\sum_j \sum_i p(c_i \mid x_i) \log(p(c_j \mid x_t)) \tag{12}$$

$$IG = H(C) - H(C \mid X) \tag{13}$$

Where $x_i$ is the possible levels in covariate X, and $c_j$ is the possible classes in target variable C. However, in regression models, the minimum sum of squared residuals (SSR) is used to reduce the variance in each of the subsets. In general, constructing a decision tree is all about successively finding the most significant attributes and splitting data in a way that returns the highest information gain (for classification) or minimum SSR (for regression). Since the measure of splitting in DT is different for discrete (information gain or IG) and continuous (variance reduction or SSR) variables, the node selection, and splitting phase becomes a multi-objective decision-making process. In other words, we construct a tree finding attributes and splitting data in such a way that maximizes IG and minimizes the SSR at the same time, as in

$$\min F_1 = \min (SSR) \tag{14}$$

$$\max F_2 = \max (IG) \tag{15}$$

These two decision-making criteria may contradict each other in some cases. For example, the attribute that increases information gain also may increase the sum of squared residuals. In this case, we may not be able to choose among some attributes since they have no priority against each other. To solve this problem, we used the non-dominated ranking approach proposed by (Deb et al., 2000). They introduced this approach to solve multi-objective optimization problems. Here, we use the same line of thinking to rank attributes based on the number of times each attribute is dominated by the other attributes regarding both splitting criteria in modeling mixed target variables. We have a set of attributes A={$x_1, x_2, \ldots, x_n$} for which we have to determine their position as a node in the tree. By definition, dominance is

$$x_i \, dom \, x_j \Leftrightarrow \forall d \;\; F_d(x_i) \le F_d(x_j) \; , \; \exists d' \; F_{d'}(x_i) < F_{d'}(x_j) \tag{16}$$

Where $d$ and $d'$ represent dimensions of decision criteria F for attributes $x_i$ and $x_j$.

**Algorithm 1:** required steps to implement MTDT

1. for each attribute in set A we calculate $F_1$ and $F_2$
2. check the dominance of attributes
3. rank attributes based on the number of times they are dominated and select Rank 1 set
4. select the one attribute from the Rank 1 set based on a secondary criterion.
5. split data based on the selected attribute
6. check the pruning criteria to avoid overfitting
7. repeat the process for each split

Although this algorithm can rank attributes based on multiple splitting criteria, still multiple attributes can be ranked similarly. To choose one of the attributes that are ranked 1, we need a secondary criterion. We introduce two approaches. 1) the modeler can give precedence to one of the criteria, i.e. either to the IG to prioritize classification or to the SSR to outweigh the regression. 2) the second approach is to select the attributes with the minimum distance to the ideal points. If there is one attribute in Rank 1, we select that one as the splitting criteria. If there are two attributes in the Rank 1 set, then we select the attributes with the closest distance to the zero points (0,0) which is the ideal point. In the case of more than two attributes in the Rank 1 set, we select the one with minimum distance to the middle point of the values of the objectives. Pseudo Code 1 gives a more systematic explanation of the entire process including the ranking algorithm and selection of attributes. We implemented this algorithm in MATLAB R2019 using the statistics and Machine learning toolbox.

**Pseudo Code 1:** Pseudo code Multi-task decision tree (MTDT)

```
A= list of attributes A={x₁,x₂, …, xₙ}
F= calculate  information gain and  the minimum sum of squared error for all xᵢ
R=Rank(A,F)
Rank1= R{1}
Best_Node = use the secondary distance criterion to choose the best attribute from the Rank 1 set.
Split  data on Best_Node and repeat the process for each split until the leaf node

Function  R=Rank (A, F)
        Sₚ=[]                          % is a list of attributes that are dominated by attribute p
        nₚ=0                           % is the number of times that attribute p is dominated by other attributes
        R{1}= []
        for p in  A
              for q in A
                    check if p dominates q
                    then add p to the Sₚ list
                    and nₚ=nₚ+1
                    otherwise
                    add q to the Sq list
                    and nq=nq+1
              end
              check if nₚ =0      % check which attributes are not dominated by any other attributes
              then R₁=[R₁  p] and Rank of p is 1
        end
        k=1
        while true
              Q=[]
              for p in R₁            % for all attribute that have rank 1
                    for q in Sₚ
                          nq=nq-1
                          check if nq =0
                          then Q=[Q  q] and Rank of q is k+1
                    end
              end
              check if Q is empty then Exit
              else
                    R_{k+1}=Q
                    k=k+1
        end
end
```

Figure 2 clearly explains how we can select between attributes with similar ranks. F1 and F2 are respectively the objective values for SSR and IG and d is the minimum distance to the middle point (i.e. $F^{mid}$) of the values of the objective functions. Assume that we have $i$ attributes that are ranked from 1 to 3. If we wanted to choose one attribute, the best would be the one with minimum $d^i$. The $F^{mid}$ and $d$ in figure 2 are computed using formulas 13 and 14, where $F_j^{max}$ and $F_j^{min}$ are the maximum and minimum of the objective $j$.

$$F_j^{mid} = \frac{F_j^{\max} - F_j^{\min}}{2} \tag{17}$$

$$d^i = \sum_{j=1}^{2} | F_j^{mid} - F_j^i |$$  (18)

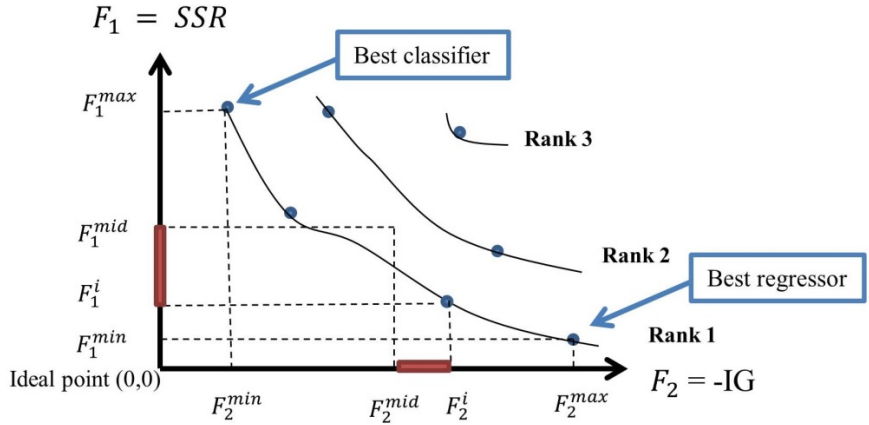

**Figure 2: Graphical representation of selecting attributes based on the secondary criterion**

We also used grid search with cross-validation to control the depth of the tree and avoid overfitting.

In summary, this algorithm is new in the following respects:

1- It can simultaneously learn two tasks i.e. classification and regression

2- It considers the correlation between discrete and continuous response variables and can predict a joint mixture

3- It uses a non-dominant sorting approach to rank and select attributes under two splitting criteria

As a measurement for the goodness-of-fit, we used the approach proposed by Arentze and Timmermans (2003) and (Kim et al., 2018). We list these measurements in Appendix A for presentation purposes.

We used this algorithm for two use cases in this paper i.e. the time-of-day model and the type-of-tour strategy model. In the next section, we present the result of the analysis based on the proposed rule-based modeling framework.

## 4    Analysis and Results

In the current section, we present the result of utilizing the zonal congestion level indicator. Secondly, we explain findings from the commodity pickup analysis and introduce 9 transport markets with their attributes. We also provide descriptive statistics of explanatory variables. Finally, we explain the results for the anatomy of tours for these transport markets through tour structure analysis.

## 4.1    Preprocessing results

### 4.1.1    Proximity to congested zones

To use the congestion level indicator defined in Equation 4 (section 3.2) in our analysis, first, we calculate the proximity of every zone to the congested zones, Because trucks may have to cross several congested zones to reach a destination that is close to or surrounded by congested zones. Proximity is defined as the Euclidean distance between the center of zones. We used the Jenks natural breaks classification method to identify different ranges of proximity to the congested PC4 zones. Figure 3 shows a heat map based on the proximity of the PC4 zones to congested zones for the morning peak period. Then we mark each zone as congested if it is close enough to a congested zone using a predefined threshold. It is important to keep this threshold small to only consider real congested zones and their surroundings. For this study, we selected the range 5653-6423 (see Figure 3) as the threshold. That is all the zones with proximity of fewer than 6423 meters to the center of congested zones are also considered congested. Decreasing this threshold to 0 means that there is no proximity considered and hence the effect of congested zones on their neighbors cannot be modeled. On the other hand, increasing the threshold would add more uncongested locations to the list of congested zones due to their proximity. This adds uncertainty to the model. The threshold, therefore, is selected in such a way that only 5% of uncongested locations will be added to the list of congested zones list due to their proximity. This way, there are relatively small to medium zones that are very close to the congested zones.

Finally, we define two binary variables identifying if the first and later pick-up or delivery locations are in a congested zone at the time of arrival. This indicator gives us information on the variation of freight activity patterns based on the perception of the congestion.

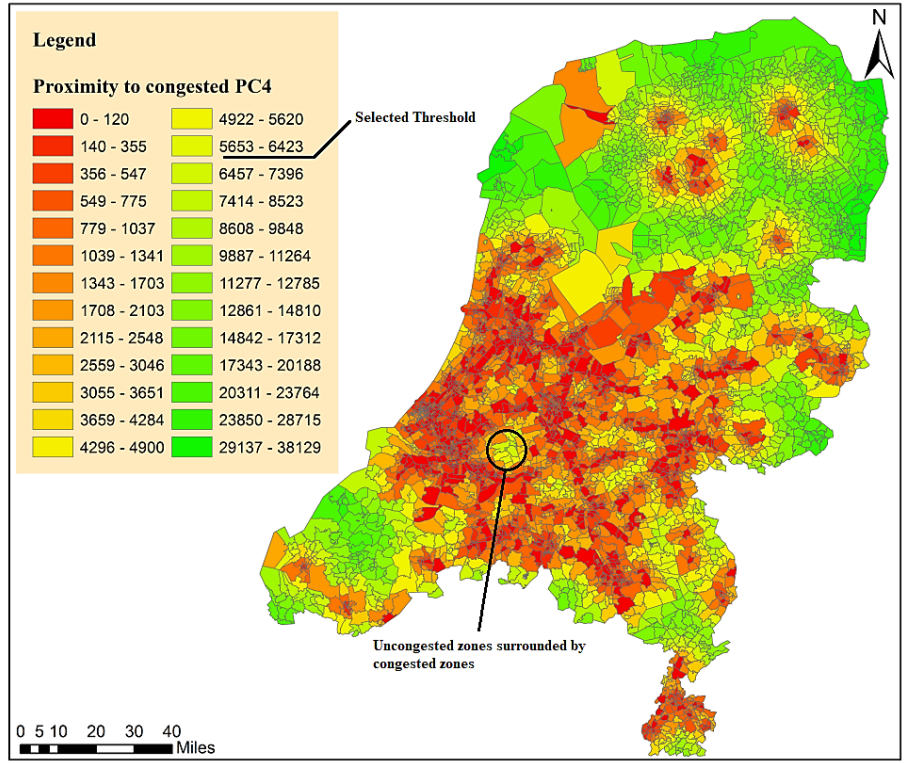

**Figure 3:** zone's cluster based on their proximity to the congested zones (PC4 - postcodes)

### 4.1.2    Transport markets and their attributes

In this section, we explain how we clustered transport markets into classes of homogeneous patterns, based on their tour activities. This leads to more accurate further modeling where specific patterns can be found inside each of these submarkets.

We use the commodity pickup analysis proposed in section 3.3.1 to identify carriers that work with multiple sectors. The minimum size of rules is set to 2 to exclude rules that identify carriers working with only one sector. Besides all these carriers that work with only one sector category (i.e. one NST-2007 category), we find 57 important rules in the diary of carriers that work with multiple sectors based on the minimum support and confidence thresholds. Figure 4 shows all these 57 rules with their support, confidence, and lift. Although all these rules are considered important, those with higher *Lift*, *Confidence*, and *Support* are the most important and certain rules.

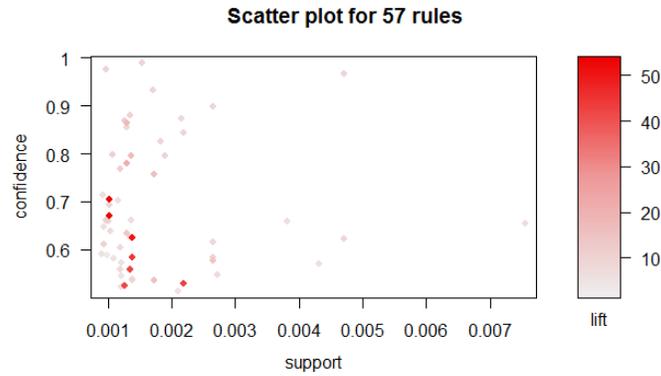

**Figure 4: Specification of extracted rules from the diary of carriers**

Figure 5 provides a graphical representation of the extracted rules with more than 0.7 confidence. There are a significant number of carriers that are active in both the food and agricultural industries. Although wood, rubber, grouped products, and fabricated metals are in different NST-R categories, we can see from the commodity pick-up analysis that these products have a high chance to be planned within the same tour of some carriers. This is because sometimes the raw material of industry belongs to a specific good category while the end product of it belongs to another class of goods. The extracted rules clearly show that carriers who are used to delivering raw materials to a firm also pick up the end products to deliver to a consumer or distribution center. The same holds for chemical and petroleum products. For example, the basic organic chemical products are the raw material for pharmaceutical industries and we can see from Figure 5 that there is a high chance that these carriers who pick up basic organic chemical products also pick up pharmaceuticals.

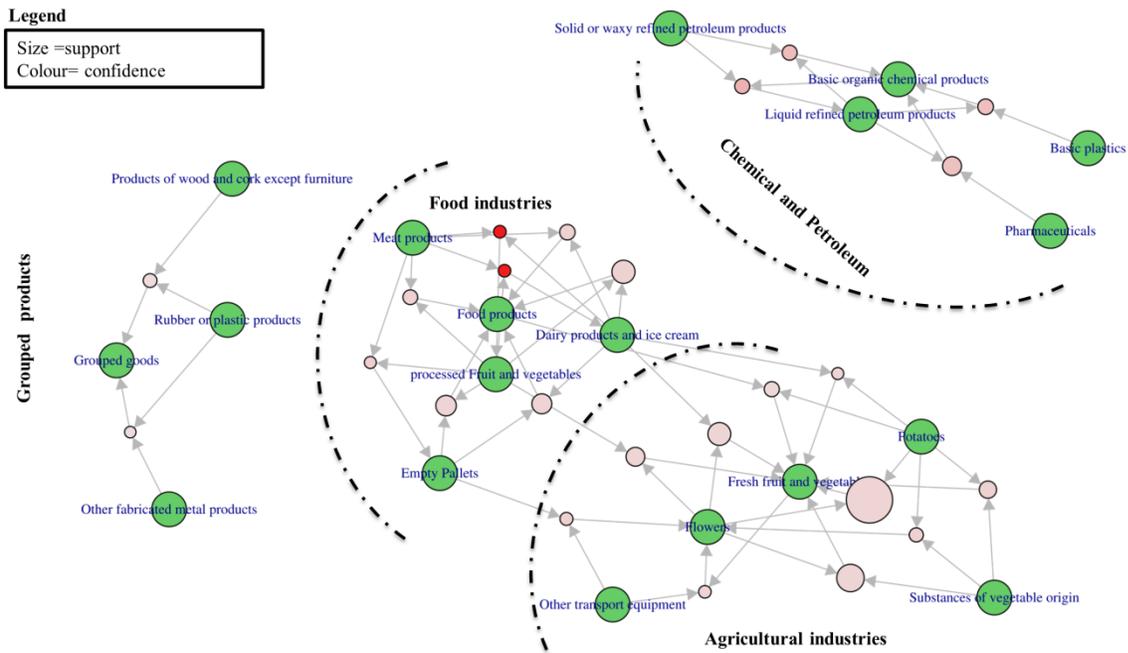

**Figure 5: Clusters of transport markets with inter-cluster and intra-cluster interactions**

In Figure 5, we can see the connections between Agricultural and Food products. This indicates how carriers connect industries between these two clusters as well as industries inside each of them. An example of the inter-cluster connection is substances of vegetable origins (e.g. seeds, stems, and roots) which are raw materials for potatoes, fresh fruit, vegetable, and flower industries. An example of an intra-cluster connection is the fact that fresh fruit and vegetables are resources for processed fruit/vegetables and food industries.

We can use the results of this algorithm to explore the probability of occurrence for all significant rules that can explain activities regarding one selected good category. For instance, Figure 6 shows the industries that produce trips for empty pallets on their tours.

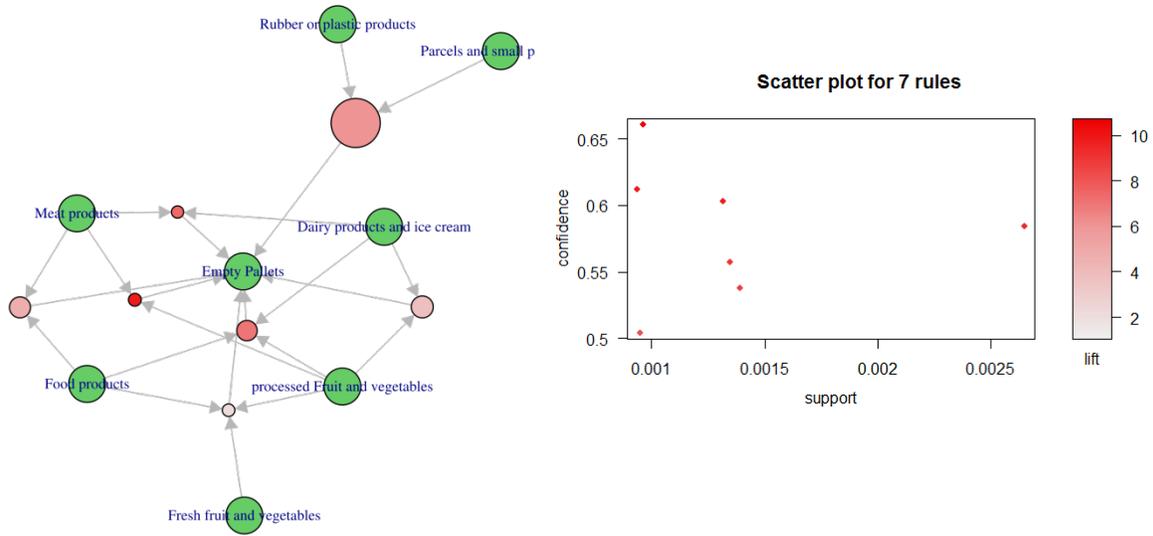

**Figure 6: Industries with a high chance to produce empty trips**

Trips with empty pallets are an important part of tours. Some industries avoid having such empty trips on their tour while it may be indispensable for other industries. From a transportation management point of view, it is important to understand which industries have a higher chance to produce such trips. Figure 6 shows 7 significant rules that indicate sectors with a higher chance of having empty trips in their tours. For instance, trips with empty pallets have a higher chance to occur in transporting parcels and rubber or plastic products. The same holds for meat products, food industries, and processed and fresh fruits and vegetables. As we can see from the above analysis, analyzing the structure of tours requires an understanding of the interaction between multiple sectors. Even carriers that work with one category of goods type, may have interactions with multiple industries within that category. Having extracted these interactions from the tour database, we can now cluster the transport market into the 9 most frequently occurring transport categories in the Netherlands. For each of these, we can then analyze the structure of tours of carriers. Table 1 shows the frequency of each of these submarkets in the database after data cleaning and pre-processing.

**Table 1: Number of observations in each transport market.**

| Transport markets | Number of observation |
| --- | --- |
| Total number of tours | 16891 |
| Tours with fresh fruit and vegetables | 3304 |
| Tours with Flowers and live plants | 772 |
| Tours with Agricultural Products | 3768 |
| Tours with Food industries | 1172 |
| Tours with Agricultural-food | 1290 |
| Tours with Chemical and petroleum | 1977 |
| Tours with Construction materials | 1038 |
| Tours with Parcels & small packages | 1693 |
| Tours with Miscellaneous goods | 1877 |

## 4.2   Explanatory variables

Table 2 indicates the sets of explanatory and response variables. They are classified into five categories including temporal variables, spatial variables, shipment characteristics, vehicle characteristics and operational attributes.

**Table 2: List of variables and their description**

| Variable | Type of variable | Description |
| --- | --- | --- |
| **Temporal variables** | | |
| Day of week | Explanatory | 0:Monday, 1:Tuesday, 2: Wednesday, 3: Thursday, 4: Friday, 5: Saturday, 6: Sunday |
| Departure time | Response | 1: Morning (6:13-10:38), 2: Midday [10:38-14:53), 3:Afternoon [14:53-19:32), 4: Night [19:32-6:13) |
| **Spatial variables** | | |
| Visit distribution centers | Explanatory | 1: if any distribution center is visited within a tour, 0: otherwise |
| Visit transshipment terminals | Explanatory | 1: if any transshipment terminal is visited within a tour, 0: otherwise |
| Congestion status of first intermediate stop | Explanatory | 1: if the visited location is marked as congested, 0: otherwise |
| Congestion status of other intermediate stops | Explanatory | 1: if the visited location is marked as congested, 0: otherwise |
| **Shipment characteristics** | | |
| Weight factor (Initial load of vehicles) | Explanatory | (0,1] |
| The number of commodities | Explanatory | Min=1, max=825 |
| Empty container/pallets | Explanatory | 1: if the tour includes trips with empty pallets or empty container |
| **Vehicle characteristics** | | |
| Vehicle Type | Explanatory | 0: truck, 1 Trailer |
| **Trip/tour operational attributes** | | |
| The average tour length | Explanatory | Min= 0, max ≈150 |
| Type of Tour | Response | Direct, collection, distribution |
| Number of stops | Response | Min=1 Max= 33 |

The temporal attributes include the day of the week as an explanatory variable and the departure time of tours as a response variable. The departure time of the tour is a continuous variable reported in minutes (0 to 1440). To keep the model tractable and reasonably simple, we categorized this variable into four clusters. We used the distribution of observed departure times to systematically come up with these categories. For this, we used a density-based clustering technique that converts a continuous variable into factors using the frequency and/or density of the variables (for more information, see the discretize function in the arule package in R programming language  (Hahsler et al., 2021)). This way, the categories follow the observed pattern in the distribution of the data. This method suggested five categories:

1- [0,373) minutes which is equal to [0:00, 6:13), labeled as ***Night***

2- [373,638) minutes which is equivalent to [6:13,10:38), labeled as ***Morning***

3- [638,893) minutes which is equivalent to [10:38,14:53) ,labeled as ***Midday***

4- [893,1172) minutes which is equivalent to [14:53,19:32), labeled as ***Afternoon***,

5- [1172,1439] minutes which is equivalent to [19:32,23:59] labeled as ***Night***.

For better presentation, we label these categorized times of the day departure time as Night, Morning, Midday, and Afternoon. Note that, we combined the first and the fifth category as we have very few observations in these categories and they both belong to the night or early morning transport when it is outside of the regular working hours.

The spatial variables relate to the characteristics and land use of the visited location in a tour. This includes the activity type of the visited firm and the congestion state of the visited zone where a visited firm is located in. The visited firm could have three types of activity i.e. transshipment terminal, distribution center, or a producer/consumer of the goods. Dummy variables are created from the activity types of the visited firms. The state of the congestion for the first and intermediate visiting locations are also coded as binary variables. In section 3.2.2, we explained how the congestion state of each zone is calculated for each hour of the day. To make use of this variable, we calculated the arrival time to each visited zone based on the observed departure time of the tour assuming that trucks took the shortest path between two pairs of zones. The status of each visited zone is considered as 1 if the zone is congested at the time of arrival and 0 otherwise.

The shipment characteristics relate to the size of the shipment. Shipment size can be explained by three aspects i.e. dimensions, weight, and the number of commodities. Among these three, the dimensions of the commodities are not observable to us from the available data. Therefore, we only used the last two for our analysis. We normalized the weight of shipments initially loaded on the vehicle based on the maximum weight of similar commodities transported in a market.

$$WF = \frac{w_{\max} - w}{w_{\max}} \tag{19}$$

This is a factor between (0,1] where the shipments with a weight factor (WF) closer to 0 are considered heavy weighted. This variable only accounts for the shipments loaded in the first loading location in a tour. Note that we are not modeling the trip sequences in a tour in this study. Therefore, the weight of the shipment in all the pickup and delivery locations cannot be a factor in this analysis. Amongst all these visiting locations, however, the weight factor of the shipment in the first loading location can play a significant role in identifying the type of tours and the number of stops.

The number of commodities varies between 1 and 825 depending on different transport markets. We also considered shipments that are aimed at transporting empty containers or pallets to depots. This variable indicates whether or not such shipment is included in a tour.

We considered vehicle type as a candidate explanatory variable for <u>vehicle characteristics</u>. Trucks and trailers are two reported vehicle types in the tour data. We created a dummy variable based on the type of vehicle.

<u>Tour/trip operational attributes</u> relate to the operational characteristics of the trucking companies. We define one explanatory and two response variables in this category. The explanatory variable is the average tour length and the response variables are the type of tours and number of stops respectively.

Shipments represent the flow of goods between loading and unloading locations. In practice, shipments are transported with commercial vehicles in an optimum way. This makes vehicle operations more complex as compared to direct shipments between loading and unloading locations. Vehicle operations consist of trips and tours. Trips are the movement of a commercial vehicle between two logistic nodes and a tour is a sequence of multiple trips. In previous research (Ruan et al., 2012, Alho et al., 2019), tours were classified into various kinds including (1) direct; (2) distribution, and (3) collection tours. Figure 7 shows the relationships between shipments and vehicle flows.

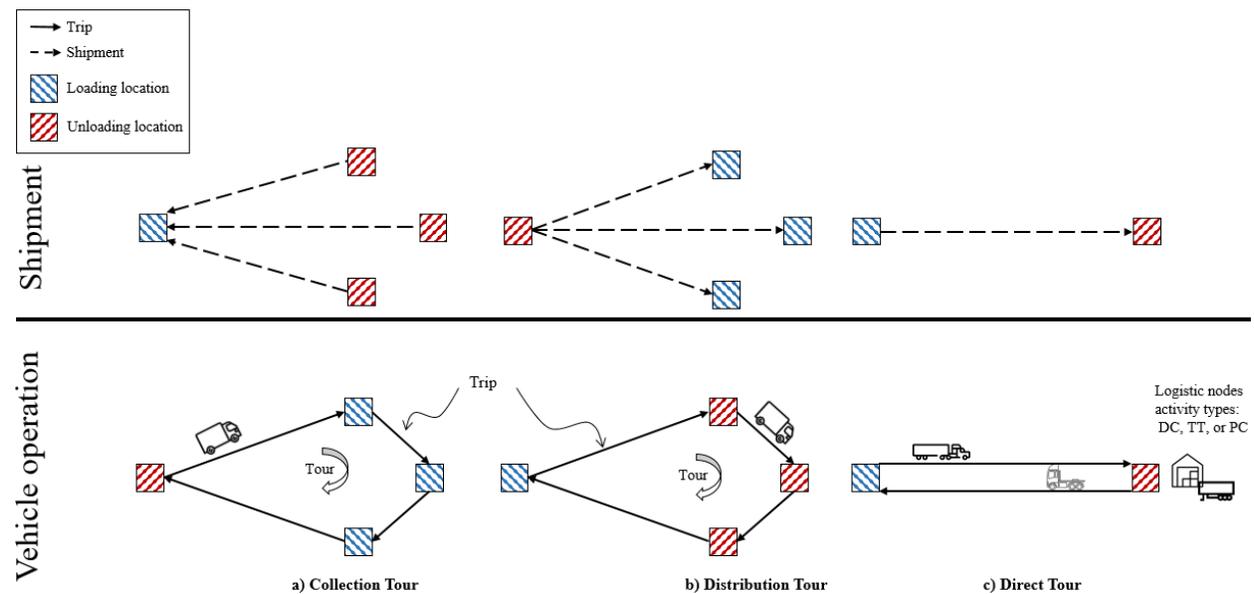

**Figure 7: Shipments and vehicle operations for 3 types of tour**

Tours are of different sizes and their size can be related to the number of stops/visits that occur in a tour. The number of stops varies between 1 and 33 depending on the transport market. The average tour length is a variable that captures the relative distance of a group of consumers to the start location of a tour. This indicates how far or close a set of customers are to the carrier. This variable varies depending on different tour type categories, different trucking companies, and different transport markets and distinguishes between local and long-haul transport patterns. Please note that this variable is not the average trip length for an observed tour since this quantity is not clear to the planners before planning the tour. Instead, this is a simple inference-based location of the customers that indicates the relative spatial relationship between each carrier and its customers. For example, some companies have longer average tour lengths in each of the tour type categories and some other companies serve more local customers.

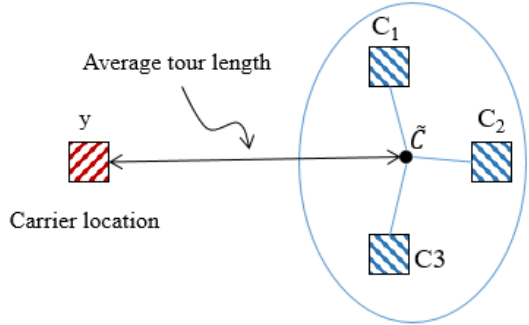

**Figure 8: Presentation of the average tour length**

Tour length is the distance between the carrier and a hypothetical costumer $\tilde{C}$ representative of all the customers that have to be visited on a tour. The $\tilde{C}$ is located in the geometric median of the customers. That is a place that has a minimum sum of the distance to all the customers ( see figure 8).

$$\arg\min_{\tilde{c}} \sum_{i=1}^{m} \| c_i - \tilde{c} \|_2 \tag{20}$$

$$Average\,tour\,length = \| \tilde{c} - y \|_2 \tag{21}$$

Given the above variable descriptions, we propose a method to analyze the structure of tours for each of these submarkets in the next section.

### 4.3   Extracted knowledge from tour structures

In this section, first, we evaluate the performance of two decision tree models, developed to study the structure of tours in freight activity of the 9 freight transport markets segmented as described above. Then, we discuss the rules identified from these two models. The first model explains the patterns in departure times of tours, while the other model provides an understanding of the identification of the type of tours. Additionally, this latter model simultaneously predicts the average number of stops per tour type using the MTDT approach proposed in section 3.3.

In this study, we used a random sample of 80% of the data to estimate the models and the rest to testing the (predictive) performance of the models. For the training set, we used 10-fold cross-validation with a 10% validation set to control the depth of the trees as well as the bias-variance trade-off. Table 3 provides the performance and accuracy of the selected models with the global commodity type as the root node. The metrics are the average of 100 runs of the fitted trees on the test data.

As we can see from this table, Cohen's kappa statistics are within the substantial agreement range (0.6-0.8) for both models. The lower Macro-F1 than Micro-F1 score suggests that the class distributions are slightly imbalanced for both models. This is a common problem for time-of-day models since the departure time of vehicles are not equally distributed over different time slots. However, we used Synthetic Minority Over-sampling Technique to reduce this effect as much as possible. The one-vs-all average accuracy (72 %) for the time-of-day model is relatively acceptable according to our application domain. The weighted random guess (WRG) accuracy indicates that both models perform far better than a model with predictions by chance. The results show the goodness-of-fit $\rho$=0.6 for time-of-day and $\rho$=0.78 for the type-of-tour model. Both models suggest a relatively high improvement in goodness-of-fit compared to that of the root models.

Table 3:Models performance and accuracy for the out-of-sample validation set

|  | Macro-F1 | Micro-F1 | One-vs-all | kappa | WRG | $\rho$ | Proot | $\rho$incr | R2 |
|---|---|---|---|---|---|---|---|---|---|
| Time-of-day model | 0.57 | 0.65 | 0.72 | 0.79 | 0.28 | 0.60 | 0.23 | 0.48 | - |
| Type-of-tour model | 0.73 | 0.81 | 0.92 | 0.68 | 0.41 | 0.78 | 0.55 | 0.51 | 0.71 |

Among the extracted rules with the two decision tree models, we can seek answers to questions like 1) how do external conditions affect peak-hour tour occurrence? 2) how do different logistics activities influence departure time and tour structure? 3) how do transport costs determine the structure of tours? 4) how does the number of commodities affects the scheduling of tours? and 5) how does vehicle type affect routing and scheduling of shipments? In order to demonstrate how these questions can be addressed, we first discuss one example in detail: food industries and agricultural products. Next, the results for other transport markets are presented.

Agricultural freight consists of products from farmlands to shops, markets and exports to other countries. Such freight encompasses a wide range of perishable goods such as vegetables, fruits, potatoes, tomatoes, cereals, livestock, raw milk, honey, and fish. The transportation of agricultural goods often faces challenges as it requires tight time schedules, high capacity and specialized equipment for loading, unloading and keeping products fresh. On the other hand, the quality, quantity and safety of food products rely on the efficiency of the food distribution system. This requires specialization not only in processing but also in transporting food products. For example, the food transportation market has adopted the functionality of keeping food fresh for an extended period through refrigerated containers.

Agricultural farming and food industries are part of the food supply chain system which ranges from seeds, fertilizers, pesticides, farming, processing, and distribution, to satisfy the demand for food. Therefore, understanding the significance of food supply chains requires comprehensive modeling since they include much more than a simple transport consideration, and involve coordinated activities. We focus here on movements from agricultural farming to the food industry. We build DT models to understand the main time-of-day patterns and tour patterns independently. Figure 9 and Figure 10 present the trees. Each DT model is composed of nodes, which are numbered from the root, e.g. the top-node 1 in Figure 9, via internal nodes to the so-called leaf nodes (the bottom row of nodes: 3, 5, 8, 9, etc.). Every leaf node consists of the probability distribution of target variables. This tree reveals the rules related to congestion, logistics nodes, transport costs, number of commodities per tour and empty trips, and type of vehicle used. As such it helps to answer our main questions. We discuss these in the remainder of this section.

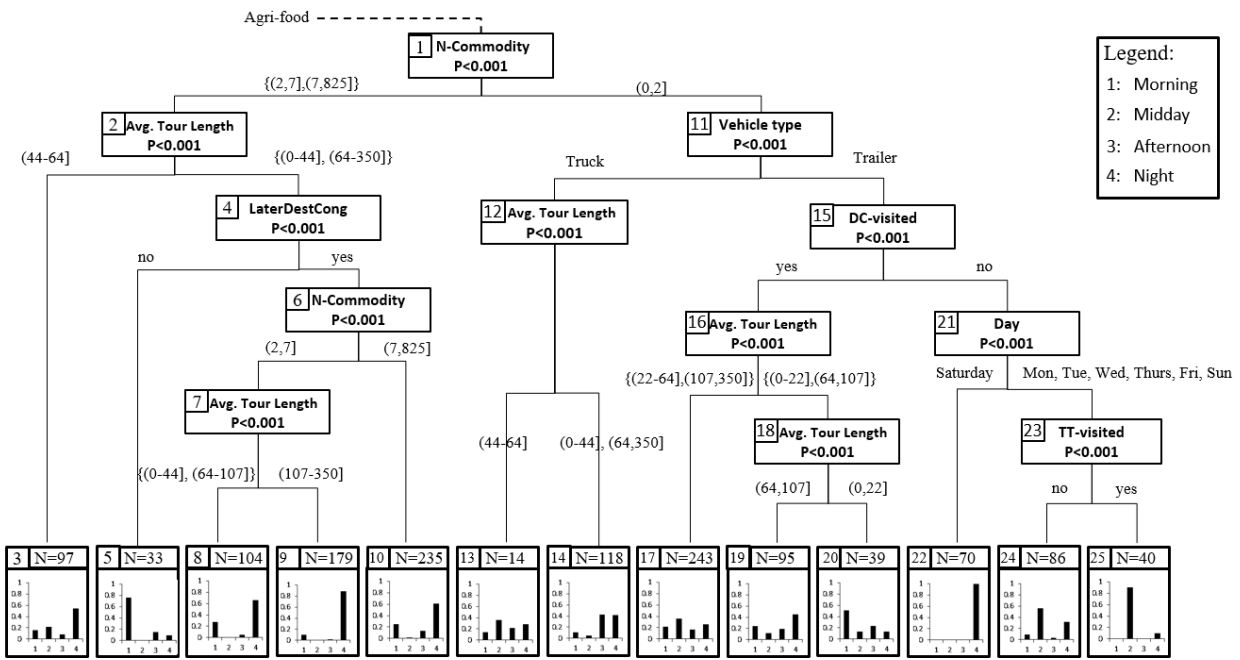

**Figure 9: Estimated tree structure for time-of-day modeling of Agricultural–Food transport.**

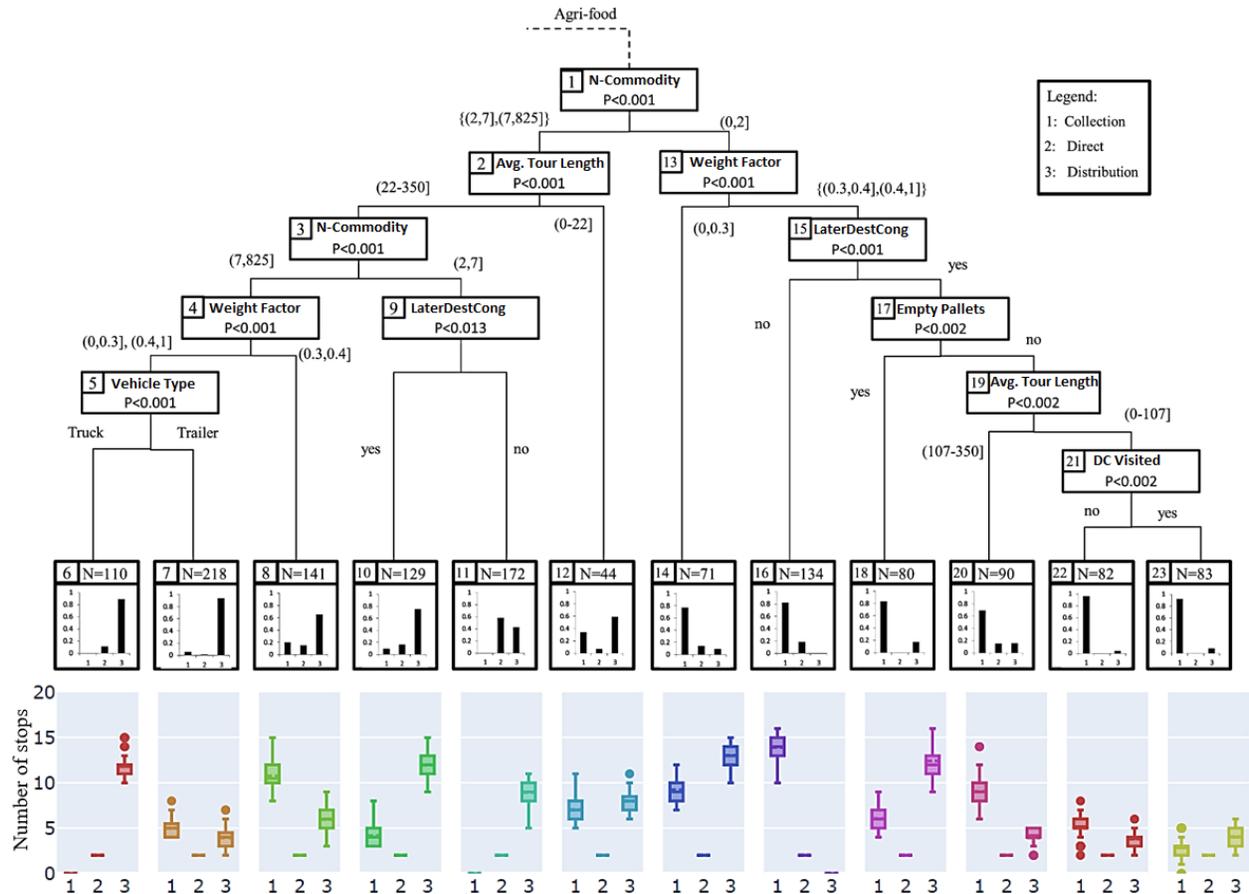

**Figure 10:Estimated tree structure for type-of-tour modeling of Agricultural-Food transport.**

### Spatial variables: Congestion-related Rules

To comprehend the time of day strategies in this market, we look at nodes related to the congestion indicator of the first and later intermediate stops. Both models indicate that the congestion state of the first pickup or delivery locations does not significantly change the departure time of tours in this market. Significant rules that identify the structure of tours facing later congested zones in their tour are:

- Comparing leaf node 5 with leaf nodes 8, 9, and 10 in Figure 9 indicates that the chance of scheduling a tour before morning peak hours (before 6:00 AM) increases if visiting pick-up or delivery locations are located in congested zones.

- In general, planners in this market schedule direct or collection tours (depending on the number of commodities) more often if there is no congestion (nodes 11 and 16, Figure 9). However, customers are served in distribution-type tours if they are located in congested zones (node 10, Figure 9).

### Spatial variables: Logistic Nodes' specific rules

In this study, we considered three types of logistics nodes that are often visited within scheduled tours: distribution centers (DC), transshipment terminals (TT), and producers/consumers. From model 1, we obtain two rules regarding the departure time of tours visiting distribution centers or terminals (Figure 9, nodes: 17, 19, 20, 25).

- If DC is nearby (less than 22 km) →Tours have a higher chance to depart during the morning peak
- If DC is far away (more than 22 km) → Tours have a higher chance to depart off-peak (Midday or Night)
- If TT visited during a tour → Tours have a higher probability to depart off-peaks (Midday or Night)

The second model does not show any significant rule for transshipment terminals. This is because of the diversity in types of tours for the export and import of agricultural-food products. It does identify tours visiting DC as ones that are usually planned in a collection tour, with 3 stops on average.

**Shipment characteristics**

One of the most important features to predict time-of-day and type of tour activities is the shipment characteristics. In this study, shipment characteristics are translated into the number of commodities, weight factor, and transportation of empty pallets/containers. The following rules are obtained:

- From Figure 9, nodes 1 and 6, we can see that planners generally plan tours with more commodities in the off-peak periods. One possible explanation for this rule is that the loading time is higher in this case and they must be scheduled in a way to avoid the peak periods and arrive on time. on the contrary, the chance of peak period departure time increases if the number of commodities decreases.

- Tours between food manufacturers and agricultural producers are often planned in either distribution or collection type. However, the probability of planning direct tours between them slightly increases for loads of lighter weights (comparing nodes 6 and 7 with 8 and 14 with 16). We elaborated this rule further in the discussion section.

- Despite the time-of-day model, the type-of-tour model suggests that transporting empty pallets in a planned tour is more likely to take place in a collection tour with a relatively low number of stops (7 stops on average).

**Vehicle characteristics**

The type of vehicle is also one of the significant explanatory variables in both models. We have two kinds of vehicles, i.e. truck and trailer, in our sample data. The following rules are obtained from the models. Mode 1 (Figure 9):

- Although tours are often planned off-peak, trucks still have a high chance of starting a tour during the Afternoon peak depending on the Tour length.

- However, trailer combinations usually start a tour off-peak either at Midday or on Nights when there is no congestion. Planners also may plan a tour with a trailer in the morning period if the average tour length is relatively low (< 22 km).

Model 2 also indicates that planned tours with trailers have fewer stops (2 on average) than trucks ( 9 on average) in distribution tours.

**Tour/Trip operational attributes**

In this study, we used the average trip length as a measurement to capture the relative adjutancy of customers to each other and the carrier. This variable distinguishes between local and long-distance transport and its impact on the time of day and type of tour patterns. We obtain the following rules to address question 3.

- In general, Figure 9 shows that tours with long average trip lengths are more likely to depart at night. Carriers should deliver commodities during working hours. Therefore, they commence the tour at night/early morning to both avoid congestion and arrive during working hours.

- We can see from Figure 10– nodes 20, 22, and 23 that the shorter the average trip length, the shorter the number of stops in the collection type of tour. In other words, planners may serve fewer local customers in one tour in a short distance and more customers over long distances.

**All the transport markets**

Similar to this analysis for transport movements between the agricultural and food industries, we have applied the models to extract knowledge for all 9 transport markets. Although decision trees are self-explanatory and can be easily understood following the paths from the root of the tree to leaf nodes, it is of interest to evaluate the quantitative effect of each of the covariates on the response variables. This helps the researcher to get a more global understanding of the model's parameters. As rule-based systems such as decision trees are non-parametric we cannot directly capture such quantitative effects. To cope with this problem, we utilized the measures proposed by Arentze and Timmermans (2003) and used in many applications such as Kim et al. (2018). These indicators quantify the magnitude and the direction of the impact of the covariates in the predicted response variable. These measures are calculated using a confusion table derived from the prediction results of each of the models. We calculate the magnitude of the impact (MI) of the covariates on the response variables as follows:

$$MI_{vi} = g(f_{vi}, f_{vi}')  \tag{22}$$

Where g is a function that measures the distance (e.g. Chi-square, likelihood, etc.) between two tables of quantities. We used Chi-square statistics in this study for the function $g$. $f_{vi}$ is the predicted frequency table for covariate $v$ on response variable $i$ and the $f_{vi}'$ is the expected frequency table assuming that the covariates $v$ has no impact on the response variable. The overall impact of the covariate $v$ on the response variable is the sum of the impacts across all the alternatives in the response variable.

$$MI_v = \sum_i MI_{vi}  \tag{23}$$

Besides the magnitude of the impact, the direction of the impact is also important for the interpretation of the models. The measure for the direction of the impact is as,

$$DI_{vi} = \frac{\sum_{j=2}^{J} (f_{i,j} - f_{i,j-1})}{\sum_{j=2}^{J} |f_{i,j} - f_{i,j-1}|}  \tag{24}$$

In this formulation the $f_{i,j}$ is the frequency of the predicted alternative $i$ in the response variable given the j[th] alternative in covariate $v$. This measurement can take any values between -1 and 1. If one covariate has a monotonically increasing impact, the $DI_{vi}$ equals 1. It would be -1 provided that the covariate $v$ has a monotonically decreasing impact. The $DI_{vi}$ between -1 and 1 indicates that the impact of the covariate on $v$ on the alternative $i$ in the response variable is non-monotonous regarding its positive or negative sign. This measurement, however, has no meaningful interpretation for covariates that are unordered (such as weekdays (S. Kim et al., 2017). We calculated the magnitude and direction of the impact for all the significant covariates derived from time-of-day and type-of-tour models. These measures are normalized across all covariates and the complete results are presented in Tables B-1 and 2 in Appendix B. These results indicate that the principles behind the structure of tours in freight activity can largely be understood under the contexts of time of day and type of tour and number of stops. The most salient findings however are discussed here. From Figure 11, we select a number of the most noticeable findings for further discussion. weight factor turns out to be irrelevant in time-of-day patterns for all sectors. Whereas it is an important factor with a relatively high magnitude of impact on type-of-day patterns in all the sectors except flowers and construction materials. The congestion level of the visiting locations is among the top three most important variables that have a high impact on both the time of day and type of tour activities.

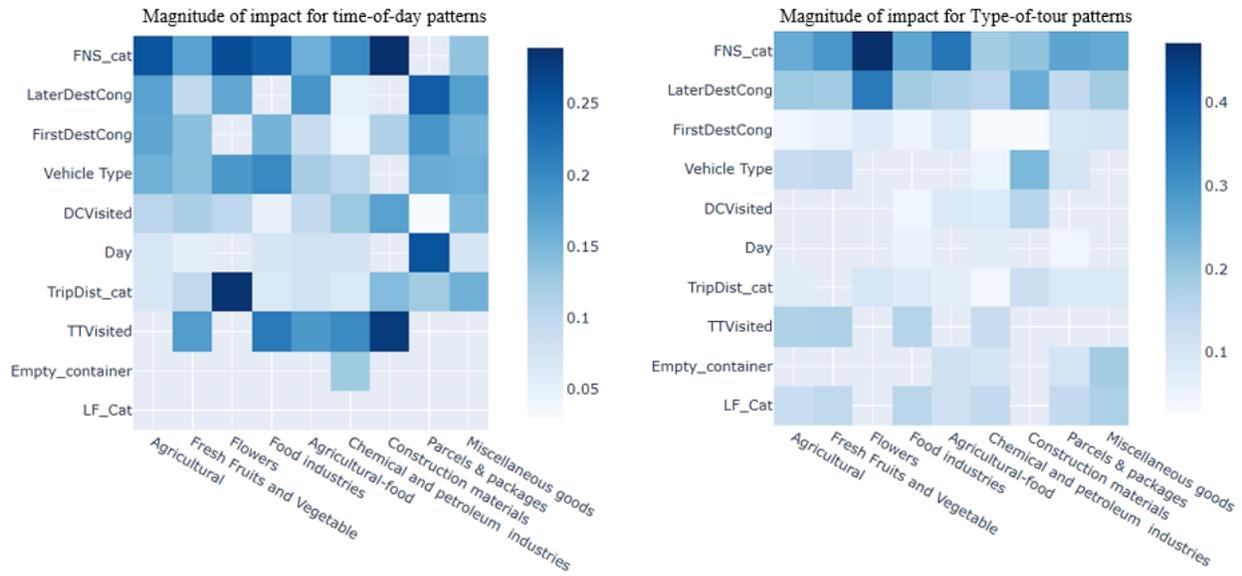

**Figure 11:The overall magnitude of the impact (*MI$_v$*) of the covariates on the response variables across all sectors.**

Regarding the direction of the impacts on time of day, the sign of impact (DI$_{vi}$), for the Agricultural, construction, and food industry transport markets (see Figure 10(a)), suggests a negative impact of visiting a congested zone on departure time slot (i.e. DI$_1$: Morning) while the impact is positive for the night/early morning departure time (i.e. DI$_4$). On contrary, the signs of impact for parcels and package markets are all positive with a high magnitude for the morning and afternoon peak periods. That is the chance of departure is high for these periods even if the locations are congested. This is mainly because the parcels and packages should often be delivered before or after the working hours of households which matches the time slots DI$_1$: Morning and DI$_3$: Afternoon. Regarding the time of day patterns, visiting transshipment terminals (TT) is also one of the most important variables (see Figure 11).

Figure 10(b) compares five different transport markets concerning visiting TT. Visiting TT has a negative impact on all the time slots for food and fresh fruit and vegetables with a larger impact on morning and night departures. This implies that there is a higher chance for departure at Midday and Afternoon for these industries. However, For the chemical and petroleum, and construction materials transport markets, visiting TT has a positive impact on the time slot Afternoon.

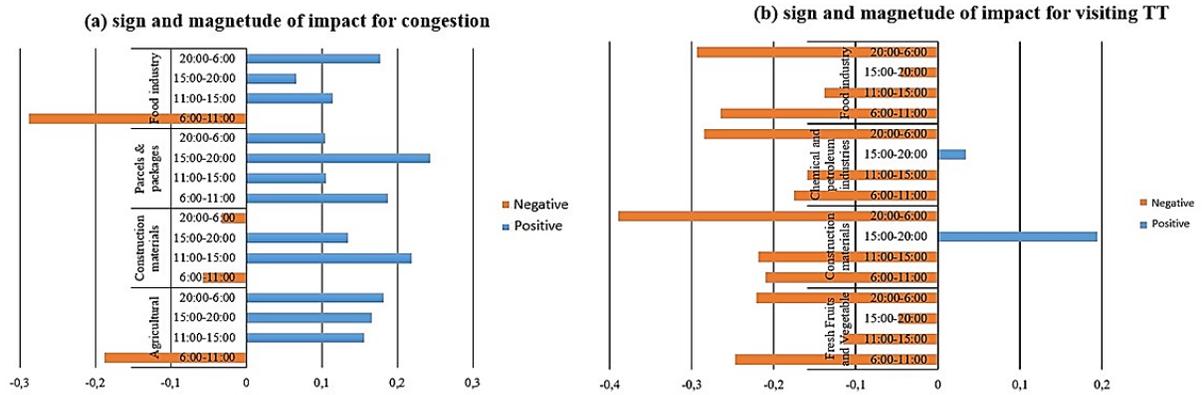

**Figure 12: Time-of-day interpretation of Sign and magnitude of impact for congestion and visiting transshipment terminals**

Regarding the impact of congestion on type-of-tour strategies, we can see from table B-2 (Appendix B) that in all markets, facing a congested location has a negative impact on direct tours. This implies that carriers often avoid planning direct tours when their customers are in congested zones. In sum, these signs and magnitudes are in line with the findings from the structure of the decision trees and our expectations. We summarize all these findings in Table 4. It provides a catalog of results for the different commodities (table rows), by summarizing the dependence structure of tours for the various contextual factors (columns), that result from the analysis. In the next section, we highlight the overall and most salient findings from the analysis.

**Table 4: Comparison of the tour characteristics and strategies for the segmented transport markets.**

| Transport market | Dealing with congestion | Influence of logistic nodes | Relation with tour length | Size of shipments | Vehicle types |
|---|---|---|---|---|---|
| Fresh Fruit and vegetables | • Planners avoid planning tours in the morning if the zone is congested.<br>• Tours are often planned off-peak.<br>• Facing congested zones→ plan in collection or distribution type of tours. | If DC visited, tours are often planned in the Afternoon.<br>If TT visited type of tour is often direct. | Larger tour length > 64 km → type of tour: direct or collection | • More shipments > 7 the higher chance for distribution type of tours.<br>• lighter weights (higher weight factor) lead to a lower probability of direct tours. | Trailers are less used for collection types of tours. |
| Flowers and live plants | • Tours depart morning even if the customers are in congested zones.<br>• Congested zones→ collection or distribution type of tour<br>• Uncongested zones → direct type of tours | No significant rule. | • Tour length > 107 → increases the chance for planning tours in early morning/night or daytime off-peak hours(midday).<br>• Tour length < 107 → plan tours in the morning even if customers are in congested zones. | • The more shipments the higher the chance for early morning/ night departure time.<br>• The weight factor is not significant neither in time of day nor in the type of tour patterns. | • Trucks are often scheduled for off-peak hours departure schemes.<br>• Trailers usually depart in the early morning except on Sundays. |
| Agricultural Products | • congested zones→ Tours depart in the early morning or at night if the average Tour length visiting locations is more than 64 km.<br>• congested zones→ tours with a lower number of commodities usually do not avoid the peak period.<br>• congested zones→ tour type: collection or distribution.<br>• not congested zone→ tour type: direct. | • departure time→<br>$\begin{cases} \text{Morning,} & \text{Tourdist} \leq 64 \text{ km} \\ \text{Afternoon,} & 64 \text{ km} < \text{Tourdist} \leq 95 \text{ km} \\ \text{Night,} & \text{Tourdist} > 95 \text{ km} \end{cases}$<br>• Tours visiting transshipment terminals are usually planned in a collection type of tour with four stops on average. | • Long tour distances → night/early morning departure.<br>• Short Tour lengths→ higher the number of stops. | • Tours with more than 2 shipments planned in the early morning or night delivery.<br>• No rule for empty trip or collection of empty pallets<br>• Tour type→<br>$\begin{cases} \text{distribution,} & N_{commodity} > 7 \\ \text{collection,} & N_{commodity} \leq 7 \end{cases}$<br>• lighter weights→ lower the probability of direct tours. | • Trucks: start a tour in the afternoon.<br>• Trailers: usually depart in the morning and also in the afternoon when there is no congestion. |
| Food industries | • congested zones→ before the morning peak departure.<br>• Congested zones → collection or distribution tours<br>• Uncongested zones → direct tours | • If DC visited → the chance for Morning departure time is higher.<br>• If TT visited → the chance for Midday departure time is higher.<br>• If TT and DC both visited → collection type of tour with a low number of stops. | • Tour length > 44 km → early morning or night departure time.<br>• Tour length < 64 km → the chance for direct tour increases<br>• Tour length > 64 km → distribution type of tour | • The higher the number of shipments the higher the chance for a morning departure time.<br>• lighter weights→ lower the probability of direct tours. | • Planners plan tours for a trailer in the early morning/night departure scheme. Whereas trucks are usually planned between 11:00 and 15:00. |
| Agricultural-food | • congested zones→ tours depart early morning.<br>• congested zones→ tour type: direct or collection<br>• not congested zone→ tour type: distribution | • DC is nearby →Tours depart during the morning peak.<br>• DC is far away → Tours depart off-peak (midday or night).<br>• TT is visited → Tours depart off-peaks (midday or night).<br>• DC has visited → the collection with 3 stops. | • Long Tour lengths → night/early morning departure.<br>• Short Tour lengths→ lower the number of stops.<br>• Long Tour lengths → more number of stops. | • Tours with more shipments are planned in the off-peak period<br>• Empty pallets are planned in collection type of tours with a lower number of stops | • Trucks: stating a tour during evening peak depending on the Tour length.<br>• Trailers: usually start a tour off-peak.<br>• Trucks: relatively higher number of stops. |

| Transport market | Dealing with congestion | Influence of logistic nodes | Relation with tour length | Size of shipments | Vehicle types |
|---|---|---|---|---|---|
| | | | | • lighter weights → higher the probability of direct tours. | |
| Chemical and petroleum products | • The probability of departure schedule for morning is high even if the vising locations are congested for the local customers.<br>• Tours are planned in off-peak early morning or night if the customer is in the congested zone and the Tour length is higher than 22 km<br>• Congested zones→ type of tour distribution or collection based on other conditions.<br>• Uncongested zones→ direct | • Planners plan in the afternoon to visit Transshipment terminals with a direct type of tour.<br>• If DC visited → for local customers (distance < 22 km) tours depart in the morning and for Tour lengths > 22 km but less than 107 km tours start at early morning/night time. | • for Tour length > 107 km→ tours starts between 11:00 and 15:00<br>• longer Tour lengths reduce the number of stops in distribution type of tours | • More than 7 shipments → morning off-peak departure.<br>• If empty container → morning off-peak departure in distribution type of tour with 2 stops on average.<br>• lighter weights→ lower the probability of direct tours. | • Trucks are more likely to depart in the morning or afternoon depending on the congestion level of the visiting zones.<br>• Trailers are more likely to depart between night or early morning.<br>• Trailers: tour type: direct<br>• Trucks: direct tours, but the chance for distribution tours increases |
| Construction materials | • pick up point is congested → the chance for departure time schedules in Midday increases.<br>• congested zones→ tour type: distribution or collection depending on the number of commodities and vehicle types<br>• not congested → direct tours | • If DC visited →The chance for morning peak departure decreases especially if the visiting locations are congested<br>• If TT visited → planners typically schedule tours at midday.<br>• If DC visited → tour type: direct with a slight increase in the chance of distribution tours. | • Long Tour lengths greater than 107 km→ morning departure.<br>• Short Tour length less than 42 → Morning departure. | • more than 7 shipments → morning off-peak departure.<br>• less than 7 shipments → type of tour: direct<br>• more than 7 shipments → type of tour: distribution<br>• The weight factor is not significant. | • No significant rules for departure time.<br>• Planners use trucks for trips with 14 stops on average whereas trailers are often used for tours with 4 stops on average. |
| Parcels & small packages | • The chance for morning or evening peak period departure is relatively high even if the visiting locations are in congested zones.<br>• Congested → Type of tour: Delivery or collection<br>• Uncongested → direct tours | If DC visited → Although parcels were transported mostly in the evening peak period, the chance for night delivery increases. | The longer the Tour length → The higher chance for night departure time. | • Empty pallets are collected in tours with 4 stops on average.<br>• The probability of collection tours increases for lighter weights. | • Trailers have a higher chance of departing during the evening peak period as compared to trucks.<br>• Trailers are used for direct and collection tours whereas trucks are used for distribution. |
| Miscellaneous goods | • Congested → although the chance for early morning departure increases, the majority of tours still depart in morning or evening peak hours.<br>• Congested → type of tour: collection or distribution<br>• Uncongested → type of tour: direct | • If DC should be visited → the chance for morning and evening peak period departure increases even if the location is congested.<br>• If DC is not visited → planners often schedule tours in midday or early morning/night period.<br>• No significant rules for the type of tour strategies. | • Long Tour lengths greater than 107 km → night/early morning departure unless DC should be visited.<br>• A short trip with less than 107 km → tours are often scheduled within the morning or evening peak period.<br>• Tours with longer than 64 km → direct tours | • The larger number of shipments (greater than 7) in the first pickup point makes dispatchers plan tours in the early morning/night scheme.<br>• Empty pallets are carried mostly in the distribution type of tours.<br>• The probability for direct tour decreases for lighter weight. | • Despite trucks, the departure time of trailers is usually planned in midday or early morning.<br>• Dispatchers plan distribution tours for trucks whereas direct tours are often planned for trailers. |

## 5  Discussion

In order to assess the ability of the proposed methods to extract meaningful and acceptable knowledge about the anatomy of tours, we summarize our main findings below and include an initial validation of outcomes against the existing literature.

The case of analyzing the structure of tours demonstrated that the new method and model proposed is capable of distilling valuable knowledge from large freight tour databases. This is the first main finding from our research. The accuracy of the proposed model is better as compared to similar approaches in the literature. For example, Khan and Machemehl (2017a) reported Rho-Square and adjusted Rho-Square of 0.56 and 0.52 respectively for the type-of-tour model using discrete-continues extreme value modeling. Whereas, the Rho and adjusted Rho for our model is 0.78 and 0.51 respectively. Similarly, for the time of day modeling reported Rho-Square and adjusted Rho-Square are 0.275 (our method reports 0.6) and 0.271 (our method reports 0.48) using the multinomial Logit model (De Jong et al., 2016). One reason for these differences is the homogeneity in tour data made by successfully applying our proposed method for the segmentation of transport markets taking tours and the activity of carriers into account (results are presented in 4.1.2).

The results of analyzing the structure of tours for all transport markets (see Table 4) shows that most of the transport markets are sensitive to congestion except carriers transporting miscellaneous goods, parcels and small packages or flowers and live plants. They still plan tours during morning peaks even if the visiting locations are congested. One possible reason for this is less demand flexibility and tight time windows in these industries. Another reason may be that these markets (especially parcels and small packages) often have high demand and hence not enough trucks to serve all demand in off-peak. This finding is in line with Holguín-Veras et al. (2008) who report that carriers transporting food, wood, metal, or construction materials are relatively responsive to off-peak period policies to avoid congestion or extra transport costs. Interestingly, our result showed that exclusive of the Agricultural-food markets, planners prefer direct tours if they face no congested zone, and they prefer distribution or collection otherwise. This result is intuitive since in most sectors direct tours are more efficient for large shipment sizes which require special facilities and more handling time for loading and unloading. In cases where customers are in congested areas, however, it would be more efficient for carriers to un consolidate shipments into a smaller size to reduce waiting time in congested areas, and instead, bundle multiple customers to maximize their capacity utilization. Nonetheless, for the Agricultural-Food markets, the evidence seems to suggest otherwise. Although Khan and Machemehl (2017a) did not consider the impact of congestion on the type of tour, they reported, in general, that distribution and collection is the most likely tour chain pattern to ship farm products and foods, which aligned well with our findings.

The Chemical and petroleum transport markets are more sensitive to congestion for local customers (e.g. gas stations) than for long-distance trips. This is because the destinations of the long-distance trips are often factories or industries with tight time windows. Carriers avoid peak hours in all industries while visiting transshipment terminals. However, the rules for visiting DCs differ from one industry to another. Visits to DCs in most of the transport markets occur during morning or afternoon peak periods, even if the locations are congested. While some markets, like fresh fruits and vegetables, have tours in the afternoon, other markets, like food industries and agricultural-food markets, plan tours in the morning. A plausible explanation is that fresh products are moved during the day before they are sold to be available for clients early in the morning, whereas food products can stay in distribution centers for a longer period. There are no freight time-of-day models in the literature that allow a systematic comparison for multiple products. The only relevant study is De Jong et al. (2016) who reported that the period 19:00 to 05:00 (night delivery) is the most preferred delivery period for wholesaler receivers on long distances. This aligns well with our finding (see table 4) that in all transport markets, except Miscellaneous goods, larger Tour lengths to visit distribution centers increase the chance for night/early morning departures.

All the transport markets are sensitive to average tour length in all dimensions (i.e. time, type, and the number of stops). Tours with a higher average tour length are more likely to depart at night or early morning. One possible reason is that carriers have to deliver the shipments to the customers during working hours. Besides, they may want to decrease travel time costs by avoiding peak period transport. The result is that local tours with short Tour lengths often have a larger number of stops. Finally, planners plan direct tours for customers with high total transport costs. As we can observe, direct tours usually have large shipment sizes or volumes. For these, the unit transport costs become low, which drives planners to arrange for a direct shipment.

The weight factor, as a shipment characteristic, is found to be insignificant for the time of the day models across all transport markets and hence is excluded from our analysis. One possible reason for this is the level of aggregation that we took into account for transport markets. Although carriers in each transport market are homogeneous, they transport goods between heterogeneous industries with different replenishment cycles and economic order quantities. In addition, our study is limited to a low resolution for the time of the day to make the model simple and tractable. For this level of resolution, the general pattern of replenishment of receivers overlaps and hence makes it difficult for the model to find a pattern. If the impact of weights on the time of day transport operations is desired, we recommend considering a lower level of resolution both in transport market segmentation and time of the day discretization.

Regardless of the time of the day, the weight factor is significant for the type of tour decisions in all transport segments except flower industries and construction materials (see Figure 10 or Table B-2 in Appendix B). For the rest of the transport segments, shipments with light weights decrease the probability of direct transport except in Agricultural-food and parcel markets. For transporting goods between a producer of agricultural products and a food manufacturer, the shipment mainly consists of light-weighted vegetables which have to be directly transported to the manufacturer to keep them fresh. On the other hand, processed food in cans are relatively heavy and has to be distributed to multiple retailers or customers. For the parcel delivery market, lighter weights would lead to a higher probability of collection tours. Postal or delivery companies like DHL usually have several pickup points where packages are received from disaggregate senders. They will plan tours to collect separate shipments with relatively low weights and deliver them to a central parcel depot, where packages will be consolidated together (shipment weight increasing because of the bundling) and after possible transport to a next hub, re-sorted and delivered to the receivers by a distribution type tour. Therefore, this is logical that lighter weights are planned in collection tours whereas heavier weights are transported more in distribution type of tours.

## 6    Conclusion

This paper investigates the anatomy of tours in freight transport and proposes a new modeling approach for understanding the complex process of transporting goods within different sectors of industry. We propose a new enhanced decision tree algorithm (Multi-Task Decision Tree) that predicts multiple discrete and continuous target variables. This algorithm can be used in transport modeling where both discrete and continuous decisions are relevant. This study also has produced new knowledge about the interaction between various markets through the analysis of the tour activities that take place between them. The measurement of these interactions allows for clustering transport markets. We explored the structure of tours in three dimensions i.e. time-of-day, tour type, and the number of stops. Based on these three features, we extracted rules from an extensive tour database to identify existing strategies for the routing and scheduling of freight vehicles, in different transport markets.

The results provide important insights into the preference of transport markets for within-peak or off-peak travel, and for the type of tour applied when facing congested zones. We showed that some markets have strict rules regarding the time of day and type of tour when visiting logistic hubs like distribution centers or transshipment terminals. All transport markets are sensitive to transport costs, as tours are more likely to depart at night or early in the morning to avoid morning peak hours. In addition, tours serving local customers with short Tour lengths often have a larger number of stops. Finally, planners tend to plan direct tours more often for customers with large distances. From a congestion avoidance perspective, planners prefer direct tours if they visit the non-congested zones and rather plan distribution or collection tours otherwise. As could be expected, these rules differ from one market to another. Our paper brings these differences and similarities into the discussion by applying a generic model to different sectors.

This research leads to the following recommendations for future work and applications in practice. From the research perspective, one can explore the application of the MTDT algorithm to model other desired mixtures of discrete and continuous target variables and capture more rules from big transport databases. Examples are the mixture of vehicle type and shipment size preferences, and the tour type and vehicle miles traveled per type of tour. we recommend also further exploration of the relationship between departure time of tours and structure of tours and number of stops since these dimensions are modeled independently in current research. This method gives a set of certain rules distilled from frequent activity patterns of transport markets but does not uncover the underlying reasons behind these rules. This topic could be investigated further through interviews with sectors or experts. Finally, transport policymakers can also benefit from this analysis, to get a better grip on the functioning of transport markets and create more effective strategies for freight demand or traffic management. From a methodological perspective, a comparative analysis between MTDT and other nonparametric machine learning methods like GAMs, random forests, and SVM on benchmark problems can provide more insights into the accuracy of the proposed MTDT method. Finally, a robust implementation of the MTDT method in open-source programming languages and statistical tools like python and R could be useful for practitioners and researchers.

### Acknowledgments

This research was supported by the Netherlands Organization for Scientific Research (NWO), TKI Dinalog, Commit2data, Port of Rotterdam, SmartPort, Portbase, TLN, Deltalinqs, Rijkswaterstaat, and TNO under project "ToGRIP-Grip on Freight Trips". The authors would like to thank Centraal Bureau voor de Statistiek (CBS) and NDW for providing us with freight tour data and traffic data, respectively. We are sincerely grateful to anonymous reviewers for all the thoughtful reviews and constructive remarks that helped us improve our paper.

## Appendix A

In this approach $\rho$ is the probability of predicting the response variable for a given observation $x$.

$$\rho = \sum_t Pr(x \rightarrow l) \sum_i \text{Pr}(i \mid l) \text{Pr}^{'}(i \mid l) \tag{25}$$

In this formulation, $Pr(x \rightarrow l)$ is the probability that the model use leaf node $l$ to classify observation $x$, $Pr(i|l)$ is the probability of the observed class $i$, and $Pr^{'}(i|l)$ is the probability of the predicted class $i$ in leaf the node $l$. It is also possible to calculate the relative performance of the models by comparing the goodness-of-fit of a model with that of its root model. The root model for a decision tree is the tree with just its root node. This measurement is comparable to the maximum likelihood ratio and indicates the model performance improvement compared to the root model.

$$\rho_0 = \sum_i \Pr(i)\Pr^{'}(i) \tag{26}$$

$$\rho_{incr} = \frac{\rho - \rho_0}{1 - \rho_0} \tag{27}$$

The most common metrics to evaluate the predictive performance of a classification tree are metrics that use precision and recall driven from the confusion matrix.

$$Accuracy(Acc) = \frac{TP + TN}{TP + TN + FP + FN} \tag{28}$$

Where TP is the number of true positives, TN is the number of true negatives, FP is the number of false-positive, and FN is the number of false negatives. However, this metric is not suitable if the class distribution is imbalanced. Therefore we also report on the one-vs-all balanced-accuracy and the F1-score which measure the performance of multi-class classifiers with the imbalanced class distribution. We also present Cohen's kappa metric which is useful to assess if the performance of the model is better than prediction by chance.

$$Sensitivity = \frac{TP}{TP + FN} \tag{29}$$

$$Specificity = \frac{TN}{TN + FP} \tag{30}$$

$$Precision = \frac{TP}{TP + FP} \tag{31}$$

$$Balanaced - Accuracy(BAcc) = \frac{1}{2}\left(Sensitivity + Specificity\right) \tag{32}$$

$$F1 - score = \frac{2 \times Sensitivity \times Precision}{Sensitivity + Precision} \tag{33}$$

As for the regression capability of the model, we report on the conventional $r^2$.

## Appendix B

**Table B-1: quantitative effect of covariates for time-of-day model**

| Market | Variables | MI | MI1 | MI2 | MI3 | MI4 | DI1 | DI2 | DI3 | DI4 |
|---|---|---|---|---|---|---|---|---|---|---|
| Agricultural | FNS_cat | 0.254 | 0.242 | 0.312 | 0.333 | 0.018 | -1 | -1 | -1 | 0.11 |
| | LaterDestCong | 0.173 | 0.011 | 0.156 | 0.166 | 0.328 | -1 | -1 | 1 | 1 |
| | FirstDestCong | 0.169 | 0.188 | 0.156 | 0.166 | 0.182 | -1 | 1 | 1 | 1 |
| | Vehtype | 0.157 | 0.217 | 0.156 | 0.166 | 0.098 | 1 | 1 | 1 | 1 |
| | DCVisited | 0.103 | 0.176 | 0.048 | 0.062 | 0.215 | 1 | 1 | 1 | 1 |
| | Day | 0.073 | 0.132 | 0.059 | 0.051 | 0.095 | 0.01 | 0.09 | 0.02 | -0.02 |
| | TripDist_cat | 0.071 | 0.034 | 0.115 | 0.055 | 0.065 | -0.12 | -0.15 | 0.34 | 0.19 |
| Fresh Fruits and Vegetable | TTVisited | 0.179 | 0.247 | 0.116 | 0.048 | 0.221 | -1 | -1 | -1 | -1 |
| | FNS_cat | 0.174 | 0.223 | 0.246 | 0.290 | 0.077 | -1 | -1 | -1 | -1 |
| | FirstDestCong | 0.140 | 0.085 | 0.123 | 0.065 | 0.192 | -1 | 1 | 1 | 1 |
| | Vehtype | 0.139 | 0.246 | 0.123 | 0.065 | 0.119 | 1 | 1 | 1 | 1 |
| | DCVisited | 0.117 | 0.039 | 0.123 | 0.025 | 0.167 | 1 | 1 | -1 | 1 |
| | LaterDestCong | 0.099 | 0.005 | 0.123 | 0.088 | 0.126 | -1 | -1 | 1 | 1 |
| | TripDist_cat | 0.097 | 0.056 | 0.101 | 0.392 | 0.048 | 0.18 | -0.12 | 0.30 | 0.06 |
| | Day | 0.055 | 0.099 | 0.044 | 0.027 | 0.050 | -0.02 | 0.11 | 0.10 | -0.03 |
| Flowers | TripDist_cat | 0.285 | 0.252 | 0.173 | 0.267 | 0.406 | 0 | -0.10 | 1 | -0.08 |
| | FNS_cat | 0.260 | 0.042 | 0.275 | 0.573 | 0.217 | -1 | -1 | -1 | -1 |
| | Vehtype | 0.185 | 0.257 | 0.137 | 0.022 | 0.268 | 1 | 1 | 1 | 1 |
| | LaterDestCong | 0.167 | 0.326 | 0.315 | 0.010 | 0.033 | 1 | -1 | 1 | -1 |
| | Day | 0.102 | 0.122 | 0.101 | 0.130 | 0.076 | -0.16 | -0.01 | 0.09 | 0.05 |
| Food industries | FNS_cat | 0.242 | 0.144 | 0.276 | 0.423 | 0.168 | 1 | -1 | -1 | -1 |
| | TTVisited | 0.216 | 0.265 | 0.138 | 0.046 | 0.294 | -1 | -1 | -1 | -1 |
| | Vehtype | 0.200 | 0.078 | 0.138 | 0.202 | 0.220 | 1 | -1 | 1 | 1 |
| | FirstDestCong | 0.153 | 0.288 | 0.115 | 0.066 | 0.178 | -1 | 1 | 1 | 1 |
| | Day | 0.074 | 0.107 | 0.072 | 0.068 | 0.072 | 0.05 | 0.67 | -0.01 | 0.00 |
| | tripDist_Cat | 0.066 | 0.053 | 0.122 | 0.147 | 0.026 | 0.17 | -0.73 | -0.42 | 0.38 |
| | DCVisited | 0.050 | 0.066 | 0.138 | 0.047 | 0.042 | 1 | -1 | 1 | 1 |
| Agricultural-food | LaterDestCong | 0.187 | 0.004 | 0.335 | 0.149 | 0.224 | -1 | 1 | -1 | 1 |
| | TTVisited | 0.185 | 0.216 | 0.062 | 0.105 | 0.243 | -1 | -1 | -1 | 1 |
| | FNS_cat | 0.158 | 0.070 | 0.200 | 0.386 | 0.012 | -1 | -1 | -1 | 1 |
| | Vehtype | 0.121 | 0.181 | 0.100 | 0.166 | 0.086 | 1 | -1 | 1 | 1 |
| | DCVisited | 0.096 | 0.048 | 0.002 | 0.020 | 0.159 | 1 | -1 | 1 | 1 |
| | FirstDestCong | 0.093 | 0.082 | 0.100 | 0.051 | 0.121 | -1 | -1 | 1 | 1 |
| | tripDist_Cat | 0.080 | 0.289 | 0.101 | 0.067 | 0.064 | -0.40 | -0.10 | 0.26 | 0.35 |
| | Day | 0.080 | 0.111 | 0.100 | 0.055 | 0.092 | 0.28 | -0.22 | -0.10 | 0.04 |
| Chemical and petroleum industries | FNS_cat | 0.200 | 0.150 | 0.328 | 0.191 | 0.128 | 1 | -1 | 0 | -0.38 |
| | TTVisited | 0.199 | 0.175 | 0.159 | 0.034 | 0.285 | -1 | -1 | 1 | -1 |
| | DCVisited | 0.129 | 0.222 | 0.154 | 0.075 | 0.033 | 1 | 1 | 1 | -1 |
| | Empty_container | 0.126 | 0.113 | 0.162 | 0.137 | 0.102 | -1 | -1 | -1 | -1 |
| | Vehtype | 0.105 | 0.025 | 0.010 | 0.210 | 0.244 | 1 | 1 | 1 | 1 |
| | Day | 0.078 | 0.105 | 0.044 | 0.104 | 0.081 | 0.07 | -0.02 | 0.17 | 0.03 |
| | tripDist_Cat | 0.066 | 0.047 | 0.079 | 0.175 | 0.054 | 0.59 | -0.24 | 0.32 | -0.06 |
| | LaterDestCong | 0.051 | 0.091 | 0.059 | 0.009 | 0.016 | 1 | 1 | 1 | 1 |
| | FirstDestCong | 0.046 | 0.073 | 0.005 | 0.065 | 0.057 | 1 | 1 | 1 | 1 |
| Construction materials | FNS_cat | 0.289 | 0.370 | 0.263 | 0.261 | 0.271 | 0.95 | -1 | 0 | -1 |
| | tripDist_Cat | 0.145 | 0.206 | 0.080 | 0.325 | 0.160 | -0.55 | -0.14 | 0.73 | -0.50 |
| | DCVisited | 0.174 | 0.157 | 0.219 | 0.086 | 0.145 | 1 | 1 | 1 | 1 |
| | TTVisited | 0.278 | 0.210 | 0.219 | 0.194 | 0.390 | -1 | -1 | 1 | 1 |
| | FirstDestCong | 0.114 | 0.057 | 0.219 | 0.135 | 0.033 | -1 | 1 | 1 | -1 |
| Parcels & packages | DCVisited | 0.030 | 0.110 | 0.048 | 0.024 | 0.000 | 1 | 1 | 1 | 1 |
| | Vehtype | 0.162 | 0.055 | 0.233 | 0.205 | 0.032 | 1 | 1 | 1 | 1 |
| | Day | 0.254 | 0.516 | 0.228 | 0.137 | 0.478 | 0 | 0.03 | -0.07 | 0.09 |
| | tripDist_Cat | 0.123 | 0.078 | 0.130 | 0.142 | 0.084 | -1 | -0.50 | 0.48 | -0.48 |
| | LaterDestCong | 0.244 | 0.055 | 0.254 | 0.248 | 0.301 | -1 | -1 | 1 | 1 |
| | FirstDestCong | 0.187 | 0.188 | 0.106 | 0.244 | 0.104 | 1 | 1 | 1 | 1 |
| Miscellaneous goods | LaterDestCong | 0.177 | 0.017 | 0.055 | 0.207 | 0.349 | -1 | -1 | 1 | -1 |
| | Vehtype | 0.159 | 0.039 | 0.166 | 0.236 | 0.111 | 1 | 1 | 1 | 1 |
| | tripDist_Cat | 0.156 | 0.226 | 0.240 | 0.150 | 0.045 | -0.75 | -0.09 | 0.45 | -0.34 |
| | DCVisited | 0.147 | 0.289 | 0.055 | 0.031 | 0.311 | -1 | 1 | 1 | 1 |
| | FNS_cat | 0.134 | 0.242 | 0.272 | 0.044 | 0.110 | -0.94 | -0.92 | 0 | -1 |
| | Day | 0.075 | 0.101 | 0.040 | 0.100 | 0.030 | 0.04 | 0.09 | -0.059 | 0.026 |
| | FirstDestCong | 0.153 | 0.086 | 0.172 | 0.232 | 0.044 | 1 | 1 | 1 | 1 |

**Table B-2: quantitative effect of covariates for type-for-tour strategies model**

| Market | variables | MI | MI1 | MI2 | MI3 | DI1 | DI2 | DI3 |
|---|---|---|---|---|---|---|---|---|
| Agricultural | FNS_cat | 0.254 | 0.356 | 0.208 | 0.251 | -1 | -1 | 0.13 |
| | LaterDestCong | 0.195 | 0.241 | 0.191 | 0.170 | 1 | -1 | 1 |
| | FirstDestCong | 0.048 | 0.084 | 0.036 | 0.039 | 1 | 1 | 1 |
| | WF_Cat | 0.126 | 0.023 | 0.189 | 0.105 | -0.60 | -0.44 | -0.75 |
| | TTVisited | 0.169 | 0.161 | 0.185 | 0.151 | -1 | -1 | -1 |
| | Vehicle Type | 0.134 | 0.087 | 0.138 | 0.161 | 1 | 1 | 1 |
| | TripDist_cat | 0.074 | 0.048 | 0.052 | 0.123 | -0.25 | -0.28 | 0.14 |
| Fresh Fruits and Vegetables | FNS_cat | 0.294 | 0.380 | 0.228 | 0.385 | -1 | -1 | 0 |
| | FirstDestCong | 0.057 | 0.090 | 0.048 | 0.050 | 1 | 1 | 1 |
| | Vehtype | 0.143 | 0.129 | 0.153 | 0.130 | 1 | 1 | 1 |
| | WF_Cat | 0.145 | 0.072 | 0.197 | 0.077 | -0.81 | -0.48 | -0.46 |
| | LaterDestCong | 0.190 | 0.190 | 0.189 | 0.192 | 1 | -1 | 1 |
| | TTVisited | 0.171 | 0.140 | 0.184 | 0.167 | -1 | -1 | -1 |
| Flowers | FirstDestCong | 0.082 | 0.160 | 0.081 | 0.061 | 1 | 1 | 1 |
| | FNS_cat | 0.471 | 0.529 | 0.423 | 0.565 | -1 | -1 | 0 |
| | LaterDestCong | 0.347 | 0.265 | 0.385 | 0.283 | 1 | 1 | 1 |
| | TripDist | 0.100 | 0.046 | 0.111 | 0.090 | -0.20 | -0.52 | 0.19 |
| Food industries | FNS_cat | 0.266 | 0.341 | 0.190 | 0.323 | -1 | -1 | 1 |
| | DCVisited | 0.044 | 0.041 | 0.012 | 0.098 | 1 | 1 | 1 |
| | LaterDestCong | 0.187 | 0.171 | 0.211 | 0.162 | 1 | -1 | 1 |
| | FirstDestCong | 0.050 | 0.039 | 0.060 | 0.045 | 1 | 1 | 1 |
| | Day | 0.055 | 0.081 | 0.041 | 0.058 | -0.16 | 0.07 | 0.03 |
| | tripDist_Cat | 0.082 | 0.120 | 0.074 | 0.065 | -0.09 | -0.64 | 0.20 |
| | WF_Cat | 0.155 | 0.124 | 0.208 | 0.097 | -1.00 | -0.52 | -0.51 |
| | TTVisited | 0.160 | 0.083 | 0.205 | 0.152 | 1 | -1 | -1 |
| Agricultural-food | LaterDestCong | 0.168 | 0.154 | 0.170 | 0.177 | 1 | -1 | 1 |
| | WF_Cat | 0.115 | 0.180 | 0.133 | 0.073 | -0.94 | 0.33 | -0.12 |
| | FNS_cat | 0.354 | 0.381 | 0.339 | 0.338 | -1 | 0 | 0.058 |
| | Vehtype | 0.090 | 0.045 | 0.029 | 0.121 | 1 | 1 | 1 |
| | DCVisited | 0.086 | 0.027 | 0.001 | 0.126 | 1 | -1 | 1 |
| | tripDist_Cat | 0.069 | 0.055 | 0.197 | 0.072 | 0.52 | 0.22 | -0.71 |
| | Empty_Pallets | 0.118 | 0.157 | 0.132 | 0.093 | -1 | -1 | -1 |
| Chemical and petroleum industries | FNS_cat | 0.187 | 0.345 | 0.091 | 0.243 | -1 | -0.89 | 1.00 |
| | WF_Cat | 0.143 | 0.067 | 0.242 | 0.062 | -0.31 | -0.71 | -0.25 |
| | DCVisited | 0.085 | 0.136 | 0.017 | 0.141 | 1 | 1 | 1 |
| | Vehtype | 0.052 | 0.032 | 0.063 | 0.045 | 1 | 1 | 1 |
| | Empty_container | 0.104 | 0.105 | 0.159 | 0.047 | -1 | -1 | -1 |
| | Day | 0.074 | 0.073 | 0.043 | 0.106 | 0.06 | -0.05 | 0.00 |
| | tripDist_Cat | 0.039 | 0.019 | 0.040 | 0.043 | 0.16 | 0.14 | 0.06 |
| | TTVisited | 0.135 | 0.148 | 0.158 | 0.109 | -1 | -1 | -1 |
| | LaterDestCong | 0.153 | 0.054 | 0.172 | 0.159 | 1 | -1 | 1 |
| | FirstDestCong | 0.029 | 0.021 | 0.015 | 0.045 | 1 | 1 | 1 |
| Construction materials | FNS_cat | 0.207 | 0.365 | 0.133 | 0.420 | -1.00 | -0.82 | 1 |
| | tripDist_Cat | 0.124 | 0.074 | 0.155 | 0.029 | 0.06 | -0.68 | 0.83 |
| | DCVisited | 0.159 | 0.051 | 0.170 | 0.170 | 1 | 1 | 1 |
| | Vehicle | 0.230 | 0.182 | 0.251 | 0.170 | 1 | 1 | 1 |
| | LaterDestCong | 0.252 | 0.182 | 0.289 | 0.142 | 1 | -1 | 1 |
| | FirstDestCong | 0.027 | 0.145 | 0.001 | 0.068 | 1 | -1 | 1 |
| Parcels & packages | Vehicle | 0.107 | 0.082 | 0.153 | 0.047 | 1 | 1 | 1 |
| | Empty_Pallets | 0.107 | 0.082 | 0.153 | 0.047 | 1 | 1 | 1 |
| | Day | 0.048 | 0.047 | 0.035 | 0.134 | 0.04 | 0.04 | -0.18 |
| | tripDist_Cat | 0.086 | 0.090 | 0.084 | 0.068 | 0.09 | -0.58 | -0.13 |
| | LaterDestCong | 0.141 | 0.166 | 0.109 | 0.134 | 1 | -1 | 1 |
| | FNS_cat | 0.268 | 0.344 | 0.164 | 0.238 | -1 | -1 | 0.03 |
| | WF_cat | 0.142 | 0.074 | 0.220 | 0.242 | 0.01 | -0.72 | -1 |
| | FirstDestCong | 0.100 | 0.115 | 0.082 | 0.089 | 1 | 1 | 1 |
| Miscellaneous goods | LaterDestCong | 0.188 | 0.166 | 0.204 | 0.204 | 1 | -1 | 1 |
| | tripDist_Cat | 0.087 | 0.090 | 0.071 | 0.201 | 0.04 | -0.82 | -0.10 |
| | FNS_cat | 0.258 | 0.331 | 0.212 | 0.163 | -1 | -1 | 0.44 |
| | Empty_Pallets | 0.188 | 0.166 | 0.204 | 0.204 | -1 | -1 | 0.03 |
| | WF_cat | 0.174 | 0.086 | 0.252 | 0.119 | -0.12 | -0.79 | -0.63 |
| | FirstDestCong | 0.104 | 0.161 | 0.057 | 0.110 | 1 | 1 | 1 |